%% file: main.tex
\documentclass{article}
\pdfoutput=1

    \PassOptionsToPackage{numbers, compress}{natbib}

     \usepackage[final]{neurips_2021}

\input{math_commands.tex}

\usepackage[utf8]{inputenc} 
\usepackage[T1]{fontenc}    
\usepackage{hyperref}       
\usepackage[capitalize]{cleveref}
\usepackage{url}            
\usepackage{booktabs}       
\usepackage{amsfonts}       
\usepackage{nicefrac}       
\usepackage{microtype}      
\usepackage{xcolor}         
\usepackage{paralist}
\usepackage{array}
\usepackage{pifont}
\usepackage{multirow}
\usepackage{amssymb}

\usepackage{graphicx}
\usepackage{algorithm}
\usepackage{algpseudocode}
\usepackage{caption}
\usepackage{subcaption}
\usepackage{adjustbox}
\usepackage{wrapfig}
\usepackage{standalone}
\usepackage{bbm} 
\usepackage{sidecap} 
\sidecaptionvpos{figure}{c} 

\definecolor{darkblue}{rgb}{0.0, 0.0, 0.55} 
\makeatletter
\def\mathcolor#1#{\@mathcolor{#1}}
\def\@mathcolor#1#2#3{%
  \protect\leavevmode
  \begingroup
    \color#1{#2}#3%
  \endgroup
}
\makeatother

\input{bib_strings}
\newcommand{\cnaps}{\textsc{CNAPs}}
\newcommand{\scl}{Simple \cnaps{} + \lite{}}
\newcommand{\sclabbrv}{SC + \lite{}}
\newcommand{\metadataset}{Meta-Dataset}
\newcommand{\vtabmd}{\textsc{VTAB+MD}}
\newcommand{\lite}{\textsc{LITE}}

\newcolumntype{P}[1]{>{\centering\arraybackslash}m{#1}}
\makeatletter
\algnewcommand{\LineComment}[1]{\Statex \hskip\ALG@thistlm \(\triangleright\) #1}
\makeatother
\usepackage{enumitem}

  \newlist{inlinelist}{enumerate*}{1}
  \setlist*[inlinelist,1]{%
          label=(\roman*),
      }

\title{Memory Efficient Meta-Learning with Large Images}

\author{%
  John Bronskill\thanks{Authors contributed equally} \\
  University of Cambridge\\
  \texttt{jfb54@cam.ac.uk} \\
  \And
  Daniela Massiceti\footnotemark[1] \\
  Microsoft Research \\
  \texttt{dmassiceti@microsoft.com} \\
  \And
  Massimiliano Patacchiola\footnotemark[1] \\
  University of Cambridge\\
  \texttt{mp2008@cam.ac.uk} \\
  \And
  Katja Hofmann \\
  Microsoft Research \\
  \texttt{kahofman@microsoft.com} \\
  \And
  Sebastian Nowozin \\
  Microsoft Research \\
  \texttt{senowoz@microsoft.com} \\
  \And
  Richard E.~Turner \\
  University of Cambridge \\
  \texttt{ret26@cam.ac.uk}
}

\begin{document}

\maketitle
\begin{abstract}
Meta learning approaches to few-shot classification are computationally efficient at test time, requiring just a few optimization steps or single forward pass to learn a new task, but they remain highly memory-intensive to train.
This limitation arises because a task's entire support set, which can contain up to 1000 images, must be processed before an optimization step can be taken.
Harnessing the performance gains offered by large images thus requires either parallelizing the meta-learner across multiple GPUs, which may not be available, or trade-offs between task and image size when memory constraints apply.
We improve on both options by proposing \lite, a general and memory efficient episodic training scheme that enables meta-training on large tasks composed of large images on a single GPU. We achieve this by observing that the gradients for a task can be decomposed into a sum of gradients over the task's training images.
This enables us to perform a forward pass on a task's entire training set but realize significant memory savings by back-propagating only a random subset of these images which we show is an unbiased approximation of the full gradient.
We use \lite{} to train meta-learners and demonstrate new state-of-the-art accuracy on the real-world ORBIT benchmark and 3 of the 4 parts of the challenging \vtabmd{} benchmark relative to leading meta-learners.
\lite{} also enables meta-learners to be competitive with transfer learning approaches but at a fraction of the test time computational cost, thus serving as a counterpoint to the recent narrative that transfer learning is all you need for few-shot classification.
\end{abstract}

\section{Introduction}
\label{sec:introduction}
Meta-learning approaches to few-shot classification are very computationally efficient.
Once meta-trained, they can learn a new task at test time with as few as 1-5 optimization steps~\citep{finn2017model,zintgraf2018cavia} or a single forward pass through the model~\citep{snell2017prototypical,requeima2019cnaps,bateni2020improved} and with minimal or no hyper-parameter tuning.
In contrast, transfer learning approaches based on fine-tuning typically rely on a large pre-trained feature extractor, and instead take 100s-1000s of optimization steps at test time in order to learn a task~\citep{kolesnikov2019big}, thus incurring a high computational cost for each new task encountered.
This makes meta-learned solutions attractive in compute-constrained deployments, or scenarios where the model must learn multiple different tasks or update on-the-fly (e.g.\ in continual and online learning settings~\citep{ring1997child, saad2009line, hoi2018online, antoniou2020defining}).

However, a crucial barrier to progress is that meta-learning approaches are memory-intensive to train and thus cannot easily leverage large images for a performance boost, as recent fine-tuning approaches have done.
This limitation arises because a meta-learner must back-propagate through \emph{all} the examples in a task's support set (i.e.\ the task's training set) that contribute to a prediction on a query example.
In some cases, this can be as many as 1000 images~\citep{dumoulin2021comparing}.
%
As a result, the amount of memory required for the computational graph grows linearly with the number of support images, and quadratically with their dimension.
In contrast, transfer learning approaches can employ standard batch processing techniques to scale to larger images when under memory constraints --- a feature which has contributed significantly to their recent success on few-shot benchmarks \citep{dumoulin2021comparing}.

Current solutions for training meta-learners on large images include 1) parallelizing the model across multiple GPUs, which may not be available or convenient, 2) considering tasks with fewer support images, and 3) employing gradient/activation checkpointing methods~\citep{chen2016training} which incur longer training times and still fall short of the task sizes required in key benchmarks.
Instead, most existing work~\citep{finn2017model,snell2017prototypical,zintgraf2018cavia,triantafillou2019meta,requeima2019cnaps,massiceti2021orbit} has opted for training on large tasks but small images which translates poorly into real-world applications and limits competitiveness on few-shot benchmarks.

\begin{SCfigure}
  \includegraphics[width=0.55\textwidth, trim={3.0cm 0.0cm 2.75cm 0.0cm}]{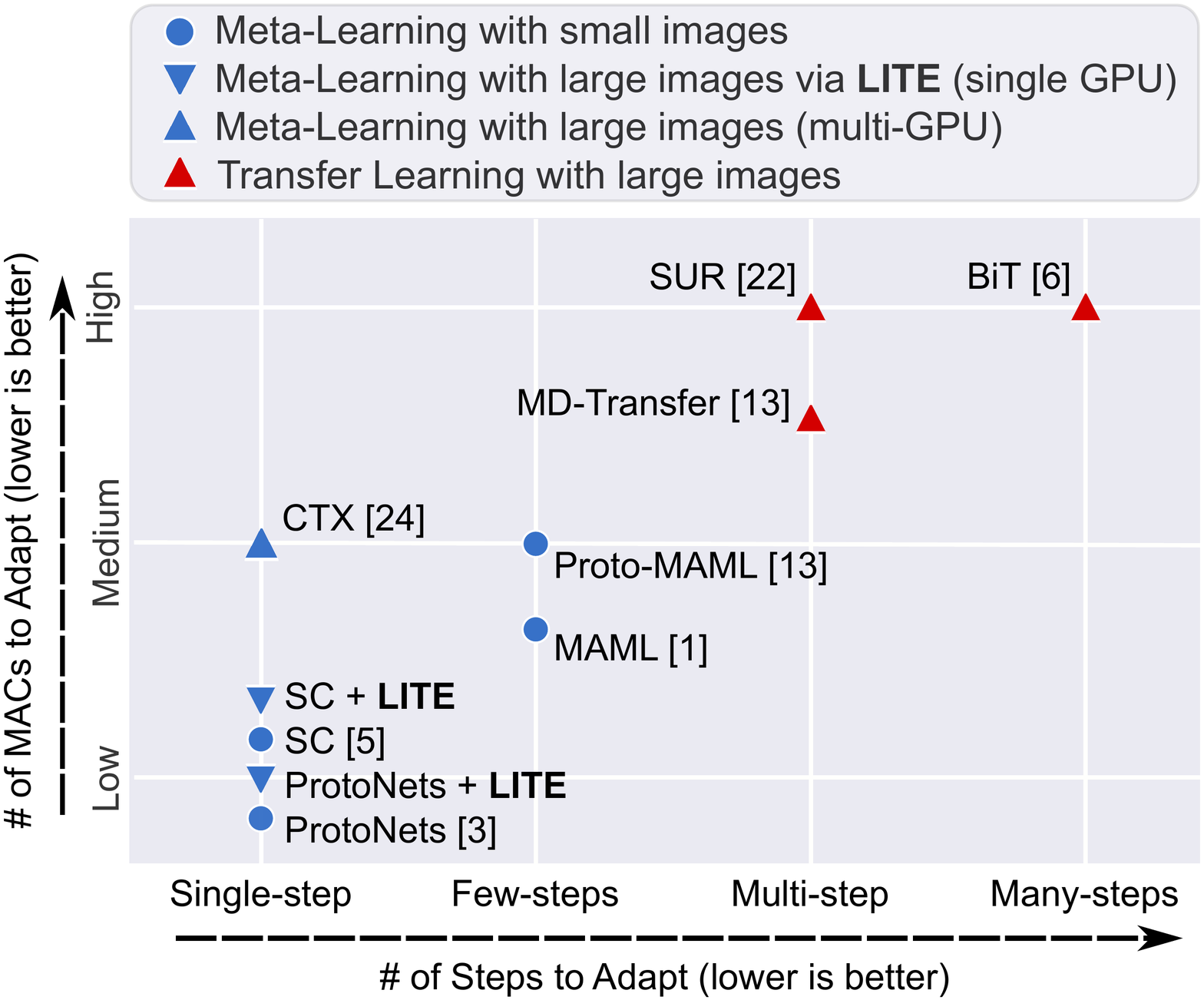} 
  \caption{\textbf{\lite{} enables meta-learners to be trained on large images with one GPU thereby significantly improving performance while retaining their test time computational efficiency.} The schematic shows test time efficiency (the number of steps and number of Multiply-Accumulate operations (MACs) needed to learn a new task at test time) and whether the method can be trained on large images (required for good performance). Existing meta-learners are cheap to adapt but trained on small images (multiple GPUs are required for large images), transfer learning methods are expensive to adapt but trainable on large images. Meta-learners with \lite{} get the best of both worlds. Note: \sclabbrv{} is Simple \cnaps{}~\citep{bateni2020improved} trained with \lite{}.}
  \vspace{-1em}
  \label{fig:models_comparison}
\end{SCfigure}

In this work, we improve on these alternatives by proposing \lite{}, a Large Image \emph{and} Task Episodic training scheme for meta-learning models that enables training with large images and large tasks, on a single GPU.
We achieve this through the simple observation that meta-learners typically aggregate a task's support examples using a permutation invariant sum operation. This structure ensures invariance to the ordering of the support set.
Consequently, the gradients for a task can be decomposed as a sum of gradient contributions from the task's support examples.
This enables us to perform a forward pass on a task's entire support set, but realize significant savings in memory by back-propagating only a random subset of these images which we show is an unbiased approximation of the true gradient.
\cref{fig:models_comparison} illustrates the key trade-offs, with \lite{} enabling meta-learners to benefit from large images for improved classification accuracy but still remain computationally efficient at test time.

We use \lite{} to train key meta-learning methods and show that our best performing instantiation -- Simple CNAPs~\citep{bateni2020improved} with LITE -- achieves state-of-the-art results relative to all meta-learners on two challenging few-shot benchmarks: \vtabmd{} \citep{dumoulin2021comparing}, an extensive suite of both meta-learning and transfer learning tasks, and ORBIT \citep{massiceti2021orbit}, an object recognition benchmark of high-variation real-world videos. 
Our results showcase the unique advantage of meta-learning methods -- that when properly trained they can be competitive with transfer learning approaches in terms of accuracy for a fraction of the computational cost at test time --- and they serve as a counterpoint to the recent narrative that transfer learning is all you need for few-shot classification.

\textbf{Our contributions}
\begin{compactenum}
\item \lite, a general and memory-efficient episodic training scheme which enables meta-learning models to be trained on large images and large tasks on a single GPU.
\item A mathematical justification for approximating the true gradient with a random subset of a task's support examples which applies to common classes of meta-learning methods.
\item Instantiations of LITE on key classes of meta-learners to demonstrate its versatility.
\item State-of-the-art performance using Simple CNAPs with LITE compared to other leading meta-learners on two challenging few-shot benchmarks, \vtabmd{} and ORBIT
%
\footnote{Source code for ORBIT experiments is available at \url{https://github.com/microsoft/ORBIT-Dataset} and for the \vtabmd{} experiments at \url{https://github.com/cambridge-mlg/LITE}.}
\end{compactenum}

\section{Why Meta-Learning with Large Images and Tasks is Difficult}
\label{sec:model_design}


%
\paragraph{Meta-learning preliminaries}
In few-shot image classification, the goal is to recognize new classes when given only a few training (or support) images of each class.
Meta-learners typically achieve this through \textit{episodic} training~\citep{vinyals2016matching}.
Here, an episode or task $\tau$ contains a support set $\mathcal{D}_S^\tau=\{(\vx^\tau_n, y^\tau_n)\}_{n=1}^{N_\tau}$ and a query set $\mathcal{D}_Q^\tau=\{(\vx^{\tau\ast}_m, y^{\tau\ast}_m)\}_{m=1}^{M_\tau}$, where $(\vx, y)$ is an image-label pair,
$N_\tau$ is the number of (labeled) support elements given to learn the new classes, and $M_\tau$ is the number of query elements requiring predictions.
Note that in a given task, elements in $\mathcal{D}_Q^\tau$ are drawn from the same set of classes as the elements in $\mathcal{D}_S^\tau$.
For brevity we may use the shorthand $\mathcal{D}_S=\{\vx, y\}$ and $\mathcal{D}_Q=\{\vx^{\ast}, y^{\ast}\}$.

\begin{figure}[h!]
    \centering
    \includegraphics[width=0.75\linewidth]{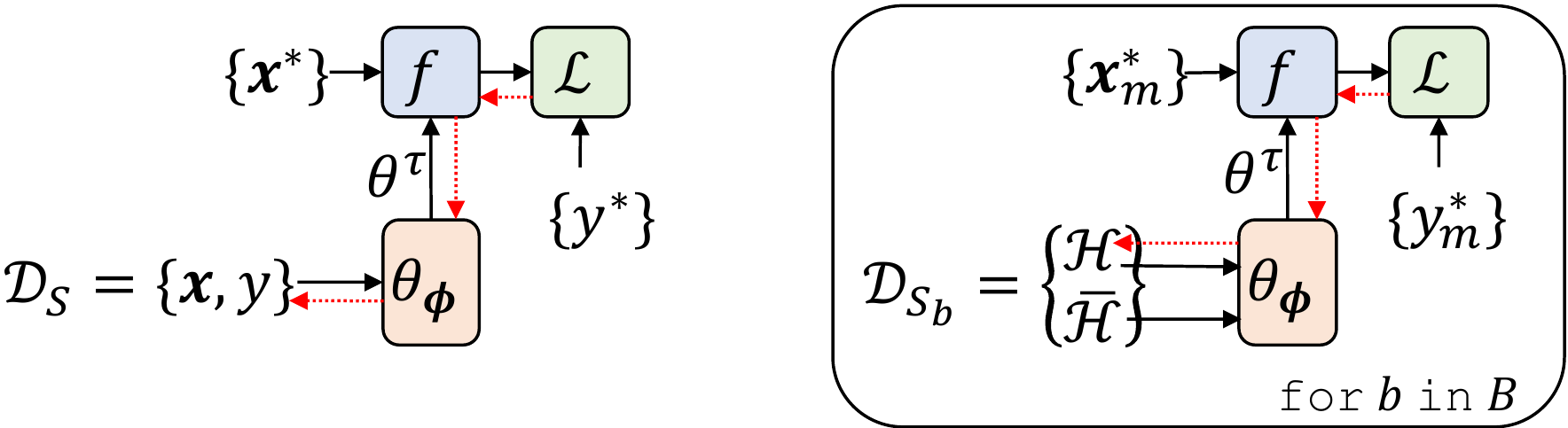}
    \caption{Left: Canonical meta-learner. Right: Meta-learner with \lite{}. The red dotted line shows the back-propagated gradients. Refer to \cref{alg:lite} for nomenclature.}
	\label{fig:lite}
	\vspace{-1.1em}
\end{figure}

During meta-training, a meta-learner is exposed to a large number of training tasks $\{\tau\}$.
For each task $\tau$, the meta-learner takes as input the support set $\mathcal{D}_S$ and outputs the parameters of a classifier that has been adapted to the current task $\vtheta^\tau = \vtheta_{\vphi}(\mathcal{D}_S)$.
The classifier can now make task-specific probabilistic predictions $f(\vx^{\ast}, \vtheta^\tau = \vtheta_{\vphi}(\mathcal{D}_S))$ for any query input $\vx^{\ast}$ (see~\cref{fig:lite}).
A function $\mathcal{L}(y^{\ast},f(\vx^{\ast}, \vtheta^\tau))$ computes the loss between the adapted classifier's predictions for the query input and the true label $y^{\ast}$ which is observed during meta-training.
Assuming that $\mathcal{L}$, $f$, and $\vtheta_{\vphi}(\mathcal{D}_S)$ are differentiable, the meta-learner can then be trained with stochastic gradient descent by back-propagating the loss and updating the parameters $\vphi$.

At meta-test time, the trained meta-learner is given a set of unseen test tasks, which typically contain classes that have \emph{not} been seen during meta-training.
For each task, the meta-learner is given its support set $\mathcal{D}_S$, and is then evaluated on its predictions for all the query inputs $\vx^\ast$ (\cref{fig:lite}, left). 

\vspace{-0.5em}
\paragraph{Large memory requirements for meta-training}
%
The primary bottleneck to using large images (i.e. $ \geq 224\times 224$ pixels) in meta-learning approaches is the large amount of (GPU) memory required to process a task's support set $\mathcal{D}_S$ during meta-training.
Specifically, the meta-learner $\theta_{\vphi}(\mathcal{D}_S)$ must perform a forward pass with a task's \emph{entire} support set before it can back-propagate the loss for query elements $(\vx^*, y) \in D_Q$ (and release the computation graph), thus preventing the use of conventional batch processing.
The amount of memory required scales linearly with the number of support images $N^\tau$ and quadratically with their dimensions.
%
If $N^\tau$ is large (e.g.~the recent \vtabmd{} benchmark \citep{dumoulin2021comparing} requires a task's support set to be as large as 1000 images), memory on a single GPU is thus quickly exceeded for large images.

Note, the number of query elements in the task $M^\tau$ is not a bottleneck when using large images as the loss decomposes over elements of the query set $\mathcal{D}_Q$ and is therefore amenable to mini-batching. By contrast, as the classifier itself is a non-linear function of the \emph{support set}, the loss does not decompose and so it is not obvious how to apply similar ideas to allow scaling of $\mathcal{D}_S$ in a principled way. 

Current ad hoc solutions to this problem are:
\begin{inlinelist}
\item parallelize the meta-learner across multiple GPUs which may not be convenient or available and can involve significant engineering effort;
\item train on tasks with smaller (or sub-sampled) support sets which may adversely affect performance on test tasks with more classes and/or large numbers of samples per class;
\item train on tasks with smaller images (e.g.~$84 \times 84$ pixels in \textit{mini}ImageNet \citep{ravi2016optimization}) which limits performance and translates poorly to many real-world applications; or
\item trade memory usage for additional computation \citep{chen2016training} by employing \textit{activation/gradient checkpointing} (i.e. during training store only a subset of intermediate activations in a network needed for backpropagation and recompute the rest with additional forward computations when needed) which allows for training on larger tasks at the expense of training time, but still falls well short of the memory needed to accommodate the task sizes required for key benchmarks (e.g. \vtabmd{}).
\end{inlinelist}

Although training meta-learners has large memory requirements, meta-testing is generally memory efficient, requiring only a small number of gradient operations, or none at all, compared to transfer learning approaches that would perform large numbers of gradient-based updates at test time.

\section{Large Image and Task Episodic (\lite{}) training}
\label{sec:lite_method}
In this section, we introduce our general and memory-efficient solution for training meta-learners episodically on tasks with large support sets and large images. We call our approach \textit{Large Image and Task Episodic} training or \lite{}.
In~\cref{sec:lite_method_applied}, we describe how \lite{} can be applied to key classes of meta-learners.
\paragraph{Approach}
The fundamental idea underlying \lite{} is to perform a forward pass using the entire support set $\mathcal{D}_S$, but to compute the gradient contribution on  only a small random subset of the examples in the support set. By doing this, we realize large savings in memory that includes gradients, activations, and the computation graph for the elements of $\mathcal{D}_S$ that are not back-propagated. This is an approximation of the true gradient that would result if back-propagation was performed on all of the examples in $\mathcal{D}_S$. In the following section, we show that this approximation is an unbiased estimate of the true gradient. The approach for a general meta-learner is detailed in \cref{alg:lite} and shown diagrammatically in \cref{fig:lite}.
\begin{algorithm}[t]
    \caption {\lite{} for a meta-training task $\tau$}
    \label{alg:lite}
    \begin{algorithmic}[1]
    \Require 
    $\mathcal{D}_S$: task support set; \ $\mathcal{D}_Q$: task query set; \ $N$: number of support examples in $\mathcal{D}_S$; \ $H$: number of elements in $\mathcal{D}_S$ to back-propagate; \ $M$: number of query examples in $\mathcal{D}_Q$; \ $M_b$: batch size for $\mathcal{D}_Q$; \  \texttt{backward()} $\equiv$ function to back-propagate a loss; \ \texttt{step()} $\equiv$ function to update parameters with a gradient step.
    \vspace{0.15cm}
    \State $B \gets \text{ceil}(M / M_b)$
    \Comment{number of query batches}
    \ForAll{$b \in 1,\dots,B$}
    \State $\mathcal{D}_{Q_b} \gets \{\vx_m^{\ast}, y_m^{\ast} \}_{m=1}^{M_b}$ \Comment{get query batch from $\mathcal{D}_Q$}
    \State $\mathcal{H} \gets \{(\vx_{n_h}, y_{n_h})\}_{h=1}^H \;\text{where}\; \{n_h\}_{h=1}^H \sim \mathcal{U}(1, N)$ \Comment{$\mathcal{H}$ to back-propagate}
    \State $\overline{\mathcal{H}} \gets \overline{\mathcal{D}_S \cap \mathcal{H}}$ \Comment{$\overline{\mathcal{H}}$ to not back-propagate}
    \State $\mathcal{D}_{S_b} \gets \mathcal{H} \cup \overline{\mathcal{H}}$
    \State $\vtheta^\tau \gets \vtheta({\mathcal{D}_{S_b}})$
    \State $L_b \gets \frac{1}{M_b}\Sigma_{m=1}^{M_b} \mathcal{L}(y_m^{\ast},f(\vx_m^{\ast}, \vtheta^\tau))$ \Comment{get loss of query batch}
    \State \texttt{backward}($L_b$) \Comment back-propagate loss on query batch
    \EndFor
    \State $\vphi \gets \texttt{step}(\vphi,N/H)$ \Comment{update $\vphi$ using weighting factor $N/H$}
    \end{algorithmic}
    \end{algorithm}

\paragraph{Mathematical justification}
The parameters of the meta-learner $\vphi$ are found by minimizing the expected loss over all tasks. 
%
%
\begin{equation}\label{eq_loss_generic}
    \underset{\vphi}{\mathrm{argmin}} \sum_{\tau=1}^T \sum_{m=1}^{M_{\tau}}
    \mathcal{L} \left( y_m^{\tau\ast}, f \left(\vx_m^{\tau\ast}, \vtheta_{\vphi}(\mathcal{D}^{\tau}_{S}) \right) \right).
\end{equation}
In most meta-learning approaches, the support set enters into the loss through a sum over the $N$ individual contributions from each data point it contains. This structure enables the meta-learners to be invariant to the ordering of the support set and allows all members of the support set to contribute to the adapted parameters (unlike alternative permutation invariant operators like \emph{max} or \emph{min}). Below we show in \textcolor{blue}{blue} how this sum arises in popular brands of meta-learners.

In amortization methods (e.g.~\textsc{CNAPs}~\citep{requeima2019cnaps} and \textsc{Versa}~\citep{gordon2018meta}), the aggregation of support set points is built in directly via a deep set encoder $e_{\vphi_{1}}(\cdot)$. This encodes the support set into an embedding vector which is mapped to the classifier parameters by a hyper-network $t_{\vphi_{0}}(\cdot)$.
\begin{equation}\label{eq_loss_amortization}
    \vtheta_{\vphi}(\mathcal{D}_{S}) = t_{\vphi_{0}}\left(\mathcolor{blue}{\sum_{n=1}^{N} e_{\vphi_{1}}(\vx_{n},y_n)} \right).
\end{equation}
In gradient-based methods (e.g.~MAML~\citep{finn2017model}), the classifier parameters are adapted from an initial value $\phi_0$ using a sum of  derivatives of an inner-loop loss computed for each data point in the support set. The derivatives play the role of the deep set encoder in amortization methods.
\begin{equation}\label{eq_loss_gradient_based}
\vtheta_{\vphi}(\mathcal{D}_{S}) = \vphi_0 + \vphi_1 \mathcolor{blue}{  \sum^{N}_{n=1} {\frac{d}{d \vphi} \mathcal{L}_{\text{inner}} (y_{n}, f(\vx_{n},{\vphi}) )}
\Bigr|_{\vphi=\vphi_0}
}
\end{equation}
Metric-based methods (e.g.~ProtoNets~\citep{snell2017prototypical}) comprise a body formed of a feature extractor and a head formed from a distance-based classifier. The classifier's body parameters are not adapted in a task specific way $\vtheta^{(\text{body})}_{\phi,c}(\mathcal{D}_{S}) = \phi_0$.  The classifier's head is adapted by averaging the activations for each class in the support set to form prototypes. Letting $k_c$ denote the number of support examples of class $c$, the adapted head parameters are given by
\begin{equation}\label{eq_loss_metric}
    \vtheta^{(\text{head})}_{\phi,c}(\mathcal{D}_{S}) = \frac{1}{k_{c}} \sum_{i=1}^{k_{c}} f(\vx_{i}^{(c)},\phi_0) = \frac{1}{k_{c}} \mathcolor{blue}{\sum_{n=1}^{N}  \mathbbm{1}(y_n=c) f(\vx_{n},\phi_0)}.
\end{equation}
Query points can then be classified using their distance from these prototypes $d(f(\vx_{n}^{\ast},\phi_0),\vtheta^{(\text{head})}_{\phi,c})$.

We have established that in many meta-learners, each support set affects the classifier parameters and therefore the loss through a sum of contributions from each of its elements. We now focus on the consequences of this structure on the gradients of the loss with respect to the meta-learner's parameters. To reduce clutter, we consider the contribution from just a single query point from a single task and suppress the dependence of the loss on the classifier and the query data point, writing
\begin{equation}
    \mathcal{L} \left( y^{\ast}, f(\vx^{\ast}, \vtheta_{\vphi}(\mathcal{D}_{S})) \right) = \mathcal{L}\left( e_{\phi}(\mathcal{D}_{S})  \right)\text{ where }\; 
    e_{\phi}(\mathcal{D}_{S}) = \mathcolor{blue}{ \sum_{n=1}^N e_{\vphi}(\vx_{n},y_n)} = \mathcolor{blue}{ \sum_{n=1}^N e^{(n)}_{\vphi} }.
\end{equation}
As a consequence of the summation, the derivative of the loss is given by 
\begin{equation}
    \frac{\mathrm{d} }{\mathrm{d} \vphi} \mathcal{L}(e_{\phi}(\mathcal{D}_{S}))
    %
    %
   =
    \mathcal{L}^{\prime}(e_{\phi}(\mathcal{D}_{S})) \times \left ( \sum_{n=1}^{N} 
    \frac{\mathrm{d}  e^{(n)}_{\vphi}}{\mathrm{d} \vphi} \right ) \text{ where } \; \mathcal{L}^{\prime}(e_{\phi}(\mathcal{D}_{S})) = \frac{\mathrm{d} \mathcal{L}(e))}{\mathrm{d} e } \Bigr|_{e=e_{\phi}(\mathcal{D}_{S})}
    \end{equation}
    which is a product of the
    sensitivity of the loss to the encoding of the data points and the sensitivity of the contribution to the encoding from each data point w.r.t.~the meta-learner's parameters. This second term is the source of the memory overhead when training meta-learners, but importantly, it can be rewritten as an expectation w.r.t.~a uniform distribution over the support set data-point indices,
 \begin{equation}
    \frac{\mathrm{d} }{\mathrm{d} \vphi} \mathcal{L}(e_{\phi}(\mathcal{D}_{S}))  = N \; \mathcal{L}^{\prime}(e_{\phi}(\mathcal{D}_{S})) \; \mathbb{E}_{n \sim \mathcal{U}(1, N)} \left[ \frac{\mathrm{d}  e^{(n)}_{\vphi}}{\mathrm{d} \vphi} \right] .
\end{equation}
We can now define the LITE estimator of the loss-derivative by approximating the expectation by Monte Carlo sampling $H$ times,
\begin{equation}
    \frac{\mathrm{d} }{\mathrm{d} \vphi} \mathcal{L}(e_{\phi}(\mathcal{D}_{S}))  \approx \frac{N}{H} \mathcal{L}^{\prime}(e_{\phi}(\mathcal{D}_{S})) \sum_{h=1}^H \frac{\mathrm{d}  e^{(n_h)}_{\vphi}}{\mathrm{d} \vphi} = \frac{\mathrm{d} }{\mathrm{d} \vphi} \hat{\mathcal{L}}(e_{\phi}(\mathcal{D}_{S})) \text{ where } \; \{n_h\}_{h=1}^H \sim \mathcal{U}(1, N) .
\end{equation}
This estimator is unbiased, converging to the true gradient as $H \rightarrow \infty$. The estimator does not simply involve subsampling of the support set -- parts of it depend on all the support set data points $\mathcal{D}_{S}$ -- and this is essential for it to be unbiased. The expectation and variance of this estimator are
\begin{equation}
\mathbb{E}_{\{ n_h \} \sim \mathcal{U}(1, N)} \left[ \frac{\mathrm{d} \hat{\mathcal{L}}}{\mathrm{d} \vphi} \right] = \frac{\mathrm{d} \mathcal{L}}{\mathrm{d} \vphi}
\; \text{ and } \; 
\mathbb{V}_{\{ n_h \} \sim \mathcal{U}(1, N)} \left[ \frac{\mathrm{d} \hat{\mathcal{L}}}{\mathrm{d} \vphi} \right] = \frac{N^{2}}{H} (\mathcal{L}^{\prime})^2 \; \mathbb{V}_{\{ n_h \} \sim \mathcal{U}(1, N)}\left[ \frac{\mathrm{d}  e^{(n_h)}_{\vphi}}{\mathrm{d} \vphi} \right]. \nonumber
\end{equation}
In \cref{sec:varying_h}, we empirically show that the \lite{} gradient estimate is unbiased and that its standard deviation is smaller than that of the naive estimator formed by sub-sampling the full support set.
\lite{} provides memory savings by subsampling $H$ examples from the support set, with $H<N$, and back-propagating only them. Crucially, a forward pass is still performed with the complementary set of points, with cardinality $N-H$, but these are not back-propagated.
\subsection{Applying \lite{} to key meta-learning approaches}
\label{sec:lite_method_applied}

%
To demonstrate its versatility, we now describe how to apply \lite{} to models within some of the main classes of meta-learners: \cnaps~\citep{requeima2019cnaps} and Simple \cnaps~\citep{bateni2020improved} for amortization-based methods and ProtoNets~\citep{snell2017prototypical} for metric-based methods.
Note, these are a few possible instantiations.
\lite{} can be applied to other meta-learning methods in a straightforward manner.

In the descriptions below, we consider just one query batch $\mathcal{D}_{Q_b}$ (i.e.\ one iteration of the for-loop in~\cref{alg:lite}).
%
Note, in practice, whenever $\mathcal{H}$ is passed through a module, back-propagation is enabled, while for $\overline{\mathcal{H}}$, back-propagation is disabled.\footnote{In PyTorch, this can be achieved by setting \texttt{\small torch.grad.enabled = True} when passing $\mathcal{H}$, and \texttt{\small torch.grad.enabled = False} when passing $\overline{\mathcal{H}}$.}
%
%
Furthermore, since typically $|\mathcal{H}| \ll |\overline{\mathcal{H}}|$, we can forward $\mathcal{H}$ in a single batch, however, we need to split $\overline{\mathcal{H}}$ into smaller batches.
Since $\overline{\mathcal{H}}$ does not require gradients to be computed, this can be done without a significant impact on memory.

\vspace{-0.5em}
\paragraph{\cnaps{}~\citep{requeima2019cnaps}, Simple \cnaps{}~\citep{bateni2020improved} + \lite{} (\cref{app:cnaps_plus_lite})}
\cnaps{} variants are amortization-based methods whose hyper-networks take a task's support set as input and generate FiLM layer~\citep{perez2018film} parameters which modulate a fixed feature extractor.
The classifier head can also be generated (\cnaps{}~\citep{requeima2019cnaps}) or adopt a metric-based approach (Simple \cnaps{}~\citep{bateni2020improved}), thus both variants can be adapted with just a single forward pass of the support set at test time.
Meta-training them with \lite{} involves passing $\mathcal{H}$ and then $\overline{\mathcal{H}}$ through their set-encoder $e_{\phi_1}$, and then averaging all the low-dimensional embeddings to get an embedding for the task.
The task embedding is then input into a set of MLPs which generate FiLM layer parameters.
$\mathcal{H}$ is passed through this configured feature extractor, followed by $\overline{\mathcal{H}}$, to get the task-adapted features for all support examples in $\mathcal{D}_{S_b}$.
For \cnaps, the task-adapted features of $\mathcal{H}$ and $\overline{\mathcal{H}}$ are pooled by class and fed into a second MLP which generates the parameters of the fully-connected classification layer.
For Simple \cnaps, the task-adapted features of $\mathcal{H}$ and $\overline{\mathcal{H}}$ are instead used to compute class-wise distributions (i.e.\ class mean and covariance matrices).
With back-propagation enabled, the query batch $\mathcal{D}_{Q_b}$ is then passed through the task-configured feature extractor and classified with the task-configured classifier (for \cnaps), or with the Mahalanobis distance~\citep{mahalanobis1936generalized} to the class-wise distributions (for Simple \cnaps).
The query batch loss is computed, and only back-propagated for $\mathcal{H}$.
Note that the feature extractor is pre-trained and frozen, and only the parameters of the set-encoder and generator MLPs are learned.
\vspace{-0.5em}
\paragraph{ProtoNets~\citep{snell2017prototypical} + \lite{} (\cref{app:protonets_plus_lite})}
ProtoNets~\citep{snell2017prototypical} is a metric-based approach which computes a set of class prototypes from the support set and then classifies query examples by their (e.g. Euclidean) distance to these prototypes.
Like \cnaps{} variants, it requires only a single forward pass to learn a new task.
Meta-training ProtoNets with \lite{} involves passing $\mathcal{H}$ through the feature extractor with back-propagation enabled, followed by $\overline{\mathcal{H}}$ with back-propagation disabled, to obtain features for all support examples $\mathcal{D}_{S_b}$.
These features are averaged by class to compute the prototypes such that (with back-propagation enabled) the query batch $\mathcal{D}_{S_b}$ can be passed through the feature extractor and classified based on the Euclidean distance.
The loss of the query batch is computed and only back-propagated for $\mathcal{H}$. 
Note that here all the parameters of the feature extractor are learned.
%
%
\section{Related work}
\label{sec:related_work}

We review the two main approaches to few-shot learning: transfer learning methods which are easy to scale to large images but are costly to adapt at test time, and meta-learning methods which are harder to scale but cheap to adapt (see~\cref{fig:models_comparison}).
Note, we do not cover methods already described above.

Transfer learning approaches have demonstrated state-of-the-art performance on challenging few-shot benchmarks~\citep{zhai2019large,dumoulin2021comparing}.
However, they incur a high computational cost at test time as they rely on large (pre-trained) feature extractors which are fine-tuned with many optimization steps.
MD-Transfer~\citep{triantafillou2019meta} fine-tunes all the parameters in a ResNet18 feature extractor with a cosine classifier head for 200 optimization steps.
BiT~\citep{kolesnikov2019big} fine-tunes a feature extractor (pre-trained on 300M images in the JFT-300M dataset~\citep{sun2017revisiting}) with a linear head, in some cases for up to 20,000 optimization steps to achieve its state-of-the-art results on the VTAB~\cite{zhai2019large} benchmark.
SUR~\citep{dvornik2020selecting} instead trains 7 ResNet-50 feature extractors, one for each training dataset.
At test time, the predictions from each are concatenated and fine-tuned to minimize a cosine-similarity loss.
All of these approaches involve on the order of teras to petas of Mulitply-Accumulate operations (MACs) to learn a single new test task, and this must be repeated for each new task encountered.
Furthermore, for each new task type, transfer learning approaches may need to be tuned on a validation set to obtain the optimal hyper-parameters.

In comparison, meta-learning approaches~\citep{hospedales2020meta} generally require orders of magnitude fewer MACs and steps to learn a new task at test time.
Popular approaches include \cnaps~\citep{requeima2019cnaps}, Simple \cnaps{}~\citep{bateni2020improved}, ProtoNets~\citep{snell2017prototypical}, and MAML~\citep{finn2017model} and are discussed in~\cref{sec:lite_method_applied}.
Others include ProtoMAML~\citep{triantafillou2019meta} which fuses ProtoNets and MAML by initializing the classifier weights with the prototypes and then, like MAML, takes a few optimization steps to tune the weights to the task.
It therefore incurs a similar cost to adapt as MAML, except it must additionally compute the prototypes.
Finally, CTX~\citep{doersch2020crosstransformers} replaces the final average pooling layer of ProtoNets with a transformer layer that generates a series of prototypes which are spatially aware and aligned with the task.
Like \cnaps, it requires just a single forward pass, however, requires more MACs to adapt since it uses a larger feature extractor (ResNet-34) and an attention module. 
CTX is one of the few meta-learning approaches that has been meta-trained on $224\times 224$ images, but requires 7 days of training on 8 GPUs.

Memory efficient variants of MAML have been developed. First-order MAML~\citep{finn2017model} saves memory by avoiding the estimate of second-order derivatives. This is also done in Reptile~\citep{nichol2018first} that additionally avoids unrolling the computation graph, performing standard gradient descent at each adaptation step. Implicit MAML~\citep{rajeswaran2019meta}, decouples the meta-gradient from the inner loop and is able to handle many gradient steps without memory constraints. In addition \cite{shin2021large}, proposes methods to reduce the computation overhead of meta-training MAML for large tasks. Note that, in all these cases, savings arise from working around the limitations of MAML, while \lite{} is more general.
%
%
\vspace{-0.5em}
\section{Experiments}
\label{sec:experiments}

In this section, we demonstrate that meta-learners trained with \lite{} achieve state-of-the-art performance among meta-learners on two challenging few-shot classification benchmarks:
\begin{inlinelist}
\item ORBIT~\citep{massiceti2021orbit} which is a real-world few-shot object recognition dataset for teachable object recognizers; and
\item \vtabmd{}~\citep{dumoulin2021comparing} which is composed of the Visual Task Adaptation Benchmark (VTAB)~\citep{zhai2019large} and Meta-Dataset (MD)~\citep{triantafillou2019meta} and combines both few-shot and transfer learning tasks.
\end{inlinelist}
We compare LITE meta-learners with state-of-the-art meta-learning and transfer learning methods in terms of classification accuracy, computational cost/time to learn a new task, and number of model parameters.

\subsection{ORBIT Teachable Object Recognition Benchmark}
\input{tables/orbit_results}

ORBIT~\citep{massiceti2021orbit} is a highly realistic few-shot video dataset collected by people who are blind/low-vision.
It presents an object recognition benchmark task which involves personalizing (i.e.~adapting) a recognizer to each individual user with just a few (support) videos they have recorded of their objects. 
To achieve this, the benchmark splits data collectors into disjoint train, validation, and test user sets along with their corresponding objects and videos.
Models are then meta-trained on the train users, and meta-tested on how well they can learn a test user's objects given just their videos (on a user-by-user basis).
The benchmark has two evaluation modes: how well the meta-trained model can recognize a test user's objects in \emph{clean} videos where there is only that object present, and in \emph{clutter} videos where that object appears within a realistic, multi-object scene.

\paragraph{Experiments} We meta-train ProtoNets~\citep{snell2017prototypical}, CNAPs~\citep{requeima2019cnaps} and Simple CNAPs~\citep{bateni2020improved} with \lite{} on tasks composed of large ($224\times224$) images.
We also meta-train first-order MAML on large images as a baseline. Since first-order MAML can process task support sets in batches, we simply reduce the batch size and do not need to use LITE.
We compare all of the above to meta-training on tasks of small ($84\times 84$) images (i.e.\ the original baselines~\citep{massiceti2021orbit}).
%
%
We also include a transfer learning approach, FineTuner~\citep{yosinski2014transferable}, which freezes a pre-trained feature extractor and fine-tunes just the linear classifier for 50 optimization steps.
For each model, we consider a ResNet-18 (RN-18) and EfficientNet-B0 (EN-B0) feature extractor, both pre-trained on ImageNet~\citep{deng2009imagenet}.
We follow the task sampling protocols described in~\citep{massiceti2021orbit} (see~\cref{app:simple_cnaps_differences,app:exp_details_for_orbit} for details).
We also include analyses on meta-training with small tasks of large images in~\cref{app:sec:small_tasks}.

\paragraph{Results}

In~\cref{tab:orbit_results}, we report frame accuracy and video accuracy, averaged over all the query videos from all tasks across all test users (17 test users, 85 tasks in total), along with their corresponding 95\% confidence intervals.
We also report the computational cost to learn a new task at test time in terms of the number of Multiply-Accumulate operations (MACs), the number of steps to adapt, and the wall clock time to adapt in seconds.
See~\cref{app:exp_details_for_orbit} for results on additional metrics.
The key observations from our results are:
\vspace{-0.5em}
\begin{itemize}[leftmargin=*]
\setlength\itemsep{0pt}
\item Training on larger ($224\times 224$) images leads to better performance compared to smaller ($84\times84$) images. The boost is significant for both clean and clutter videos, though absolute performance remains lower on clutter videos. This suggests that object detection or other attention-based mechanisms may be required to further exploit large images in more complex/cluttered scenes.
\item All meta-learners + \lite{} set a new state-of-the-art on clean videos, and perform competitively with the FineTuner on cluttered videos, using an EfficientNet-B0 backbone.
\item Meta-learners are competitive with transfer learning approaches in accuracy but are almost two orders of magnitude more efficient in the number of MACs and the time to learn a new task, and one order of magnitude smaller in the number of steps to adapt.
\end{itemize}

\subsection{\vtabmd{}}
\begin{figure}[t]
\centering
\vspace{-1.5em}
\includegraphics[width=1.0\textwidth, trim={4.10cm 0.90cm 4.10cm 1.65cm}, clip]{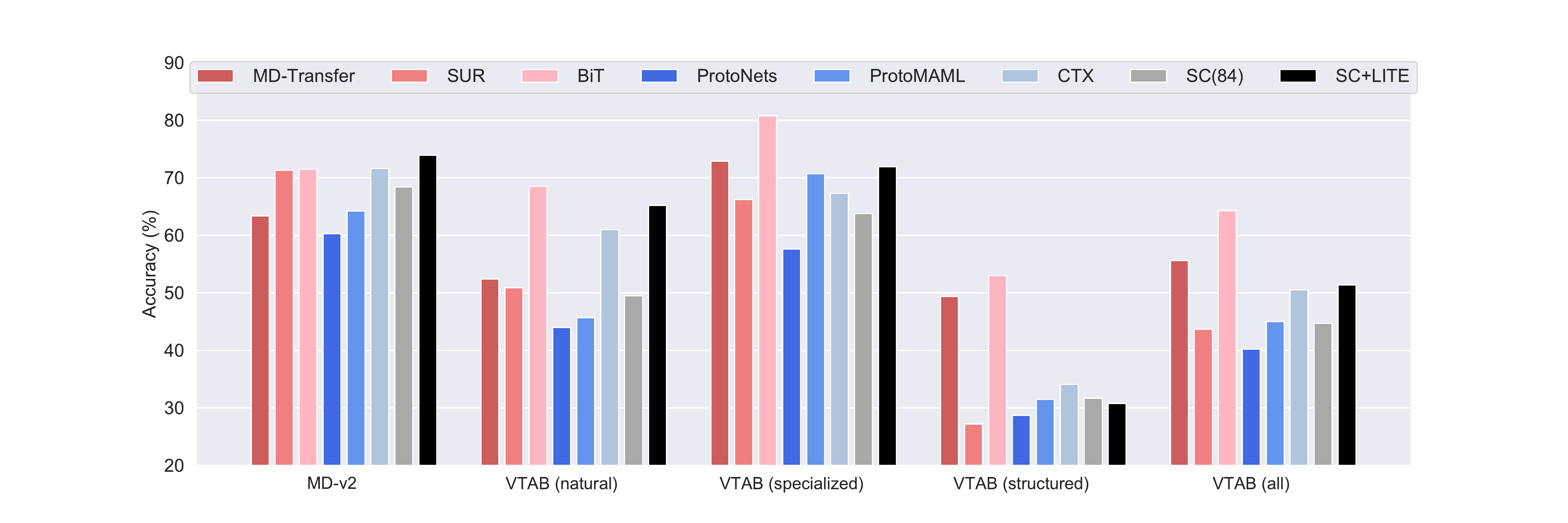} 
\caption{Summary of results on \vtabmd{}. \textbf{Simple CNAPs with LITE (\sclabbrv, black bar) trained on $224\times224$ images achieves state-of-the-art accuracy on Meta-Dataset (MD-v2), and state-of-the-art accuracy among meta-learners on 3 of 4 parts of VTAB}. As reference, we have included transfer learning methods (red bars), other meta-learning methods (blue bars), and Simple CNAPs without LITE trained on small images ($84\times84$, gray bar).  Competitive results from \citep{dumoulin2021comparing}. See \cref{tab:vtab+md_details} for tabular results on individual datasets.} 
\label{fig:vtab_md_results}
\vspace{-1.5em}
\end{figure}

\vtabmd{} \citep{dumoulin2021comparing} combines revised versions of the \metadataset{} (MD) and VTAB datasets and is one of the largest, most comprehensive, and most challenging benchmarks for few-shot learning systems.
The MD-v2 part of the benchmark involves testing on 8 diverse datasets while VTAB-v2 involves testing on 18 datasets grouped into three different cataegories (natural, specialized, and structured).
\paragraph{Results}
\cref{fig:vtab_md_results} compares our best meta-learner, Simple \cnaps{} + \lite{}, with 6 other competitive approaches on \vtabmd{}: BiT, MD-Transfer, and SUR are transfer learning based methods while ProtoMAML, ProtoNets, and CTX are meta-learning based.
Note, comparisons are not always like-for-like as methods use differing backbones, image sizes, and pre-training datasets.
%
\cref{app:sec:vtab_tabular_results} provides this information along with the precise numbers and 95$\%$ confidence intervals.
\cref{app:simple_cnaps_differences,app:exp_details_for_vtabmd} summarize all implementation and experimental details, and~\cref{app:sec:small_tasks} includes further analyses on the impact of task size on meta-training.
The key observations from our results are:
\vspace{-0.5em}
\begin{itemize}[leftmargin=*]
\setlength\itemsep{0pt}
\item On MD-v2, \scl{} has the highest average score and sets a new state-of-the-art.
\item On VTAB-v2, BiT scores highest overall, but among meta-learners, \scl{} is the best overall and on the natural and specialized sections. It falls short of CTX on the structured section due to poor performance on dSprites which involves predicting the position and orientation of small white shapes on a black background -- a task quite different to image classification. 
\item These results are significant given that CTX takes 7 days to train on 8 GPUs, whereas \scl{} trains in about 20 hours on a single 16GB GPU. In addition, \scl{} uses a relatively small pre-trained backbone (4.0M parameters) compared to SUR's 7 pre-trained ResNet-50s (one for each MD-v2 training set, plus ImageNet).
\item \scl{} using 224$\times$224 images significantly outperforms Simple \cnaps{} using 84$\times$84 images, except for when the dataset images are small (e.g.\ Omniglot, QuickDraw, dSprites).
This demonstrates that using large images and an approximation to the support set gradients achieves superior results compared to using small images with exact gradients.
\item MD-Transfer and BiT perform strongly across \vtabmd{}, however, \scl{} is significantly faster to adapt to a new task requiring only a forward pass of the support set with no hyper-parameter tuning whatsoever. The transfer learners instead perform 100s of optimization steps to adapt to a new task and may require a human-in the-loop to tune hyper-parameters such as the learning rate and number of optimization steps.
\end{itemize}
\subsection{Effect of varying $|\mathcal{H}|$}
\label{sec:varying_h}
\begin{wrapfigure}{r}{0.42\textwidth}
	\centering
    \includegraphics[width=0.42\textwidth, trim={0.25cm 0.35cm 0.4cm 0.25cm}, clip]{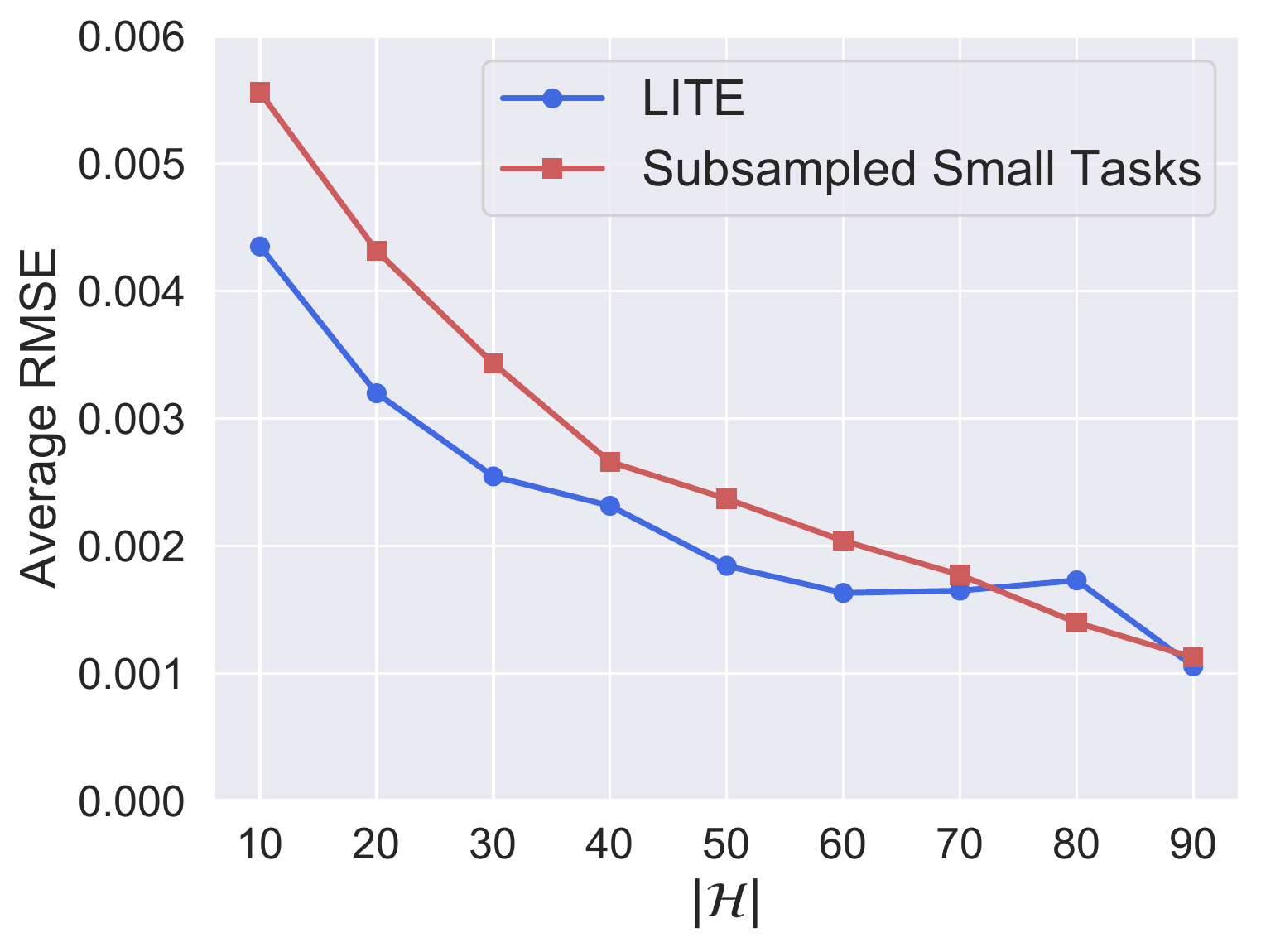} 
	\caption{Average Root Mean Square Error (RMSE) w.r.t.~the true gradients versus $|\mathcal{H}|$ for \lite{} and sub-sampled tasks on $84 \times 84$ images (10-way, 10-shot, $|\mathcal{D}_S|=100$).} 
	\label{fig:std_vs_h}
	\vspace{-1.0em}
\end{wrapfigure}
\cref{tab:vary_h_summary} shows the effect of varying $|\mathcal{H}|$, the number of examples back-propagated per task, on the VTAB+MD benchmark.
Performance is consistent across different $|\mathcal{H}|$, an expected result since \lite{} provides an unbiased estimate of the true gradient.
For both Simple CNAPs and ProtoNets + \lite{}, the results at the lowest values of $|\mathcal{H}|$ are respectable, though they fall short of what can be achieved at $|\mathcal{H}|=40$. Thus, the support gradient information does improve the solution, albeit by only 1-2 percentage points.
Note, we report the lowest setting as $|\mathcal{H}|=1$ for Simple CNAPs but $|\mathcal{H}|=0$ for ProtoNets.
This is because Simple CNAPs's adaptation network (which processes the support set) shares no parameters with the feature extractor and thus will not be learned if the support gradients are completely ignored.
On the other hand, ProtoNets' adaptation network shares all its parameters with the feature extractor, thus can be meta-learned even when the support gradients are neglected.
Finally, in the two rightmost columns, we compare the classification accuracy for $|\mathcal{H}|=|\mathcal{D}_S|$ (i.e. using the full support set gradient) to $|\mathcal{H}|=40$ (i.e. using LITE). Due to memory constraints, we do this at image size $84 \times 84$.
Here we see that the difference in accuracy is significant.
We expect that performance will smoothly interpolate as $|\mathcal{H}|$ is increased from 40 to the size of the largest support set (at which point the full gradient is computed).
This validates how LITE can be used to trade-off GPU memory usage for classification accuracy by varying $|\mathcal{H}|$.
Furthermore, we conduct an empirical analysis (see~~\cref{tab:mse_vs_h}) which shows that the \lite{} gradient estimates and the gradients when using smaller sub-sampled tasks  are unbiased w.r.t.~the true gradients.
However,~\cref{fig:std_vs_h} shows that \lite{} offers a significantly lower root mean square error w.r.t.~the true gradients compared to using sub-sampled tasks at all but the highest values of $|\mathcal{H}|$.
Refer to \cref{app:varying_h} for additional details.
%
\input{tables/varying_h_summary}

\section{Discussion}
\label{sec:discussion}

We propose \lite{}, a general and memory-efficient episodic training scheme for meta-learners that enables them to exploit large images for higher performance with limited compute resources.
\lite{}'s significant memory savings come from performing a forward pass on a task's full support set, but back-propagating only a random subset, which we show is an unbiased estimate of the full gradient.
We demonstrate that meta-learners trained with \lite{} are state-of-the-art among meta-learners on two challenging benchmarks, ORBIT and \vtabmd, and are competitive with transfer learning approaches at a fraction of the test time computational cost.

This offers a counterpoint to the recent narrative that transfer learning approaches are all you need for few-shot classification.
Both classes of approach are worthy pursuits (and will need to exploit large images in real-world deployments) but careful consideration should be given to the data and compute available at test time to determine which class is best suited to the application under consideration.
If it involves learning just a single task type (e.g. classifying natural images) with ample data and no compute or time constraints, then a fine-tuning approach would suffice and perform well.
However, if a multitude of task types will be encountered at test time, each with minimal data, and new tasks need to be learned on resource-constrained devices (e.g.\ a mobile phone or a robot) or quickly/repeatedly (e.g.\ in continual or online learning settings), then a meta-learning solution will be better suited.

Finally, as the machine learning community grapples with greener solutions for training deep neural networks, \lite{} offers a step in the right direction by allowing meta-learners to exploit large images without an accompanying increase in compute.
Future work may look toward applying the basic concept of \lite{} to other types of training algorithms to realize similar memory savings.

\paragraph{Limitations}
As discussed in~\cref{sec:lite_method}, \lite{} can be applied to a wide range of meta-learners provided that they aggregate the contributions from a task's support set via a permutation-invariant operation like a sum.
Because only a subset of the support set is back-propagated, however, the gradients can be more noisy and meta-training may require lower learning rates.
Furthermore, \lite{} is a memory-efficient scheme for training meta-learners episodically and has not been tried with meta-learners trained in other ways (e.g.~with standard supervised learning) or non-image datasets.

\paragraph{Societal impact}

Few-shot learning systems hold much positive potential -- from personalizing object recognizers for people who are blind~\citep{massiceti2021orbit} to rendering personalized avatars~\citep{zakharov2019few} (see~\citep{hospedales2020meta} for a full review).
These systems, however, also have the potential to be used in adverse ways -- for example, in few-shot recognition in military/surveillance applications.
Meta-trained few-shot systems may also pose risks in decision making applications as uncertainty calibration in meta-learning models has not yet been extensively explored.
Careful consideration of the intended application, and further study of uncertainty quantification in meta-learning approaches will be essential in order to minimize any negative societal consequences of \lite{} if deployed in real-world applications.

\section*{Acknowledgments}
We thank the anonymous reviewers for key suggestions and insightful questions that significantly improved the quality of the paper.
Additional thanks go to Vincent Dumoulin for providing the tabular results for SUR used in \cref{fig:vtab_md_results} and \cref{tab:vtab+md_details}.

\section*{Funding Transparency Statement}
Funding in direct support of this work: John Bronskill, Massimiliano Patacchiola and Richard E. Turner are supported by an EPSRC Prosperity Partnership EP/T005386/1 between the EPSRC, Microsoft Research and the University of Cambridge. 

\bibliography{references}
\bibliographystyle{unsrtnat}

\newpage
\setcounter{figure}{0}
\setcounter{table}{0}
\setcounter{equation}{0}

\appendix

\renewcommand\thefigure{\thesection.\arabic{figure}} 
\renewcommand\thetable{\thesection.\arabic{table}} 
\renewcommand\theequation{\thesection.\arabic{equation}}



%
\section{Applying \lite{} to meta-learners}

\subsection{\cnaps{}, Simple \cnaps{} + \lite}
\label{app:cnaps_plus_lite}
\begin{figure}[h!]
	\centering
    \includegraphics[width=\textwidth]{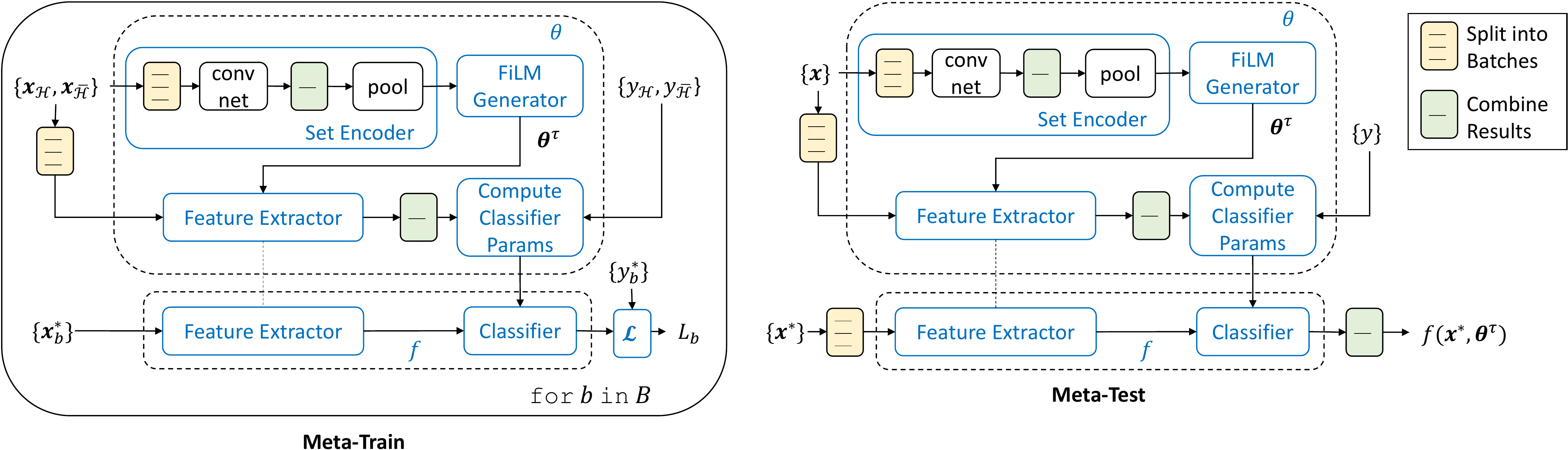}
	\caption{\cnaps{}~\citep{requeima2019cnaps}, Simple \cnaps{}~\citep{bateni2020improved} with \lite{} processing flow.}
	\label{fig:cnaps_plus_lite}
\end{figure}

We show the \lite{} processing flow for both meta-training and meta-testing phases of \cnaps{}/Simple \cnaps{} in~\cref{fig:cnaps_plus_lite}.
For each query set meta-training batch $b$, the support set images $\{ \vx_{\mathcal{H}}, \vx_{\overline{\mathcal{H}}} \}$ are broken into batches, passed through a 2D conv net, then coalesced so that the pooling step can compute the mean of all the support set embeddings.
This mean embedding is then passed into the FiLM parameter generator so that the feature extractor can be configured for the task.
The support set images $\{ \vx_{\mathcal{H}}, \vx_{\overline{\mathcal{H}}} \}$ are then passed through the adapted feature extractor in batches and the outputs are coalesced and then fed along with the support set labels $\{ y_{\mathcal{H}}, y_{\overline{\mathcal{H}}} \}$ into the box labeled "Compute Classifier Params".
For \cnaps{}, this box performs the class-conditional pooling operation and then uses an MLP to generate the weights and biases for the linear classifier.
For Simple \cnaps{}, the same box computes the class-conditional means and covariances that are then used by the classifier in the Mahalanobis distance calculations.
Once the classifier has been configured, the images in the query set batch $\{ \vx_b^* \}$ can be classified and along with the true labels $\{ y_b^* \}$, a loss is then computed.
The meta-testing flow is similar, with the exception of the loss computation.
\subsection{ProtoNets + \lite}
\label{app:protonets_plus_lite}
\begin{figure}[h!]
	\centering
    \includegraphics[width=1.0\textwidth]{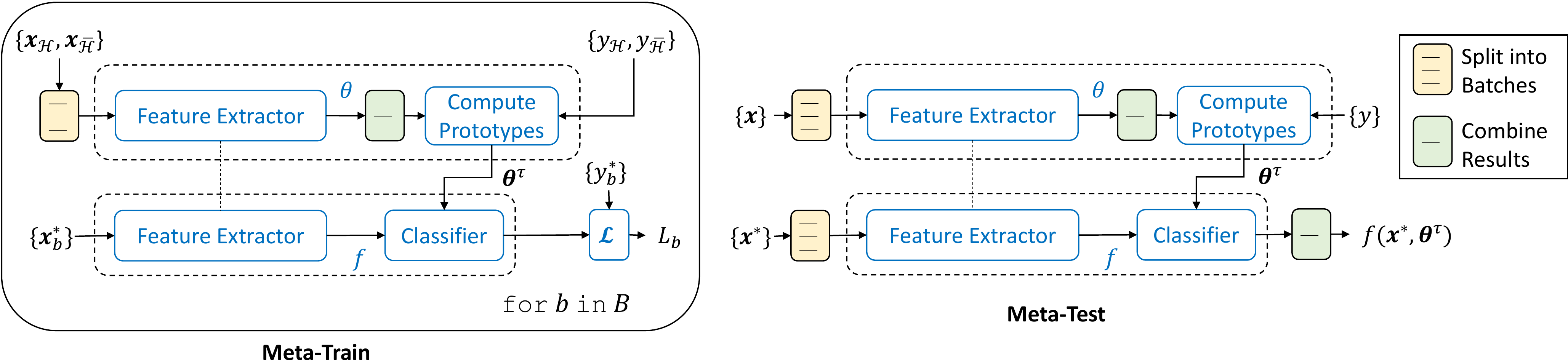}
	\caption{ProtoNets~\citep{snell2017prototypical} with \lite{} processing flow.}
	\label{fig:protonets_plus_lite}
\end{figure}

We show the \lite{} processing flow for both meta-training and meta-testing phases of ProtoNets in~\cref{fig:protonets_plus_lite}.
For each query set meta-training batch $b$, the support set images $\{ \vx_{\mathcal{H}}, \vx_{\overline{\mathcal{H}}} \}$ are broken into batches, passed through the feature extractor and the resulting embeddings are then combined.
The combined embeddings along with the support set labels $\{ y_{\mathcal{H}}, y_{\overline{\mathcal{H}}} \}$ are then used to compute the class prototypes.
The query batch images $\{ \vx_b^* \}$ are then passed through the feature extractor and the Euclidean distance from each query set image embedding to each of the class prototypes is computed.
The predicted class is the one with the minimum distance. These predictions along with the true labels $\{ y_b^* \}$ are used to compute the loss.
The meta-testing flow is similar, with the exception of the loss computation.
%
%
\section{Additional Simple \cnaps{} details}
\label{app:simple_cnaps_differences}

\begin{figure}
	\centering
    \includegraphics[width=1.0\textwidth]{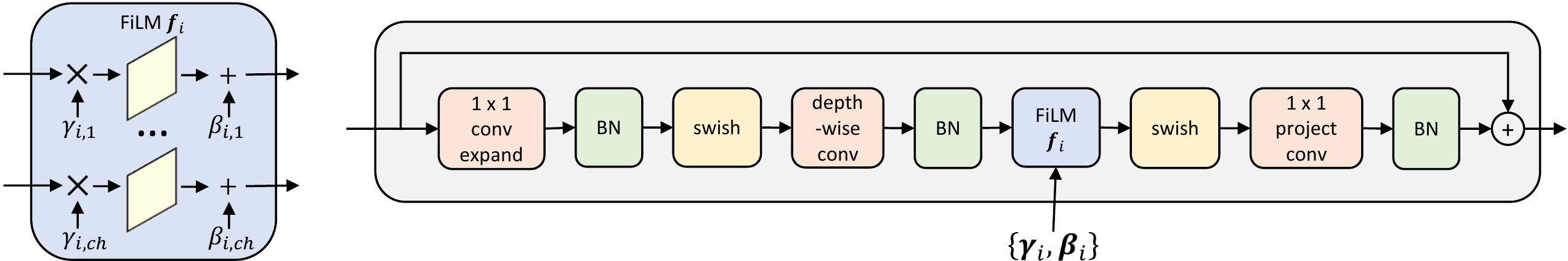}
	\caption{(Left) A FiLM layer operating on convolutional feature maps indexed by channel $ch$. (Right) How a FiLM layer is used within a inverted residual block \citep{sandler2018mobilenetv2} of an EfficientNet \citep{tan2019efficientnet}.}
	\label{fig:film}
\end{figure}

\begin{figure}
	\centering
    \includegraphics[width=0.8\textwidth]{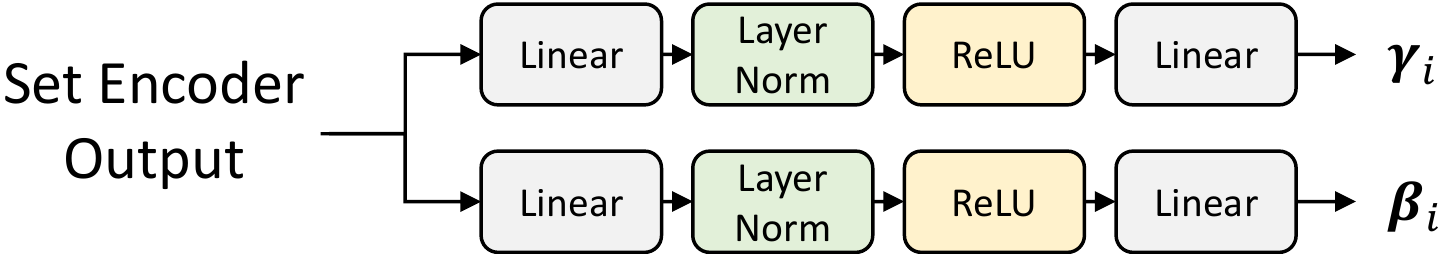}
	\caption{Generator for $i^{th}$ FiLM layer. The generator takes the output of the set encoder step for each task $\tau$ and passes it through the network to generate the parameters $\vgamma_i$ and $\vbeta_i$. The dimension of the vectors  $\vgamma_i$ and $\vbeta_i$ is equal to the number of feature channels at the location where the $i^{th}$ FiLM layer is placed within feature extractor. The depicted generator network structure is repeated for each FiLM layer added to the feature extractor.}
	\label{fig:film_generator}
\end{figure}

Our implementation of Simple \cnaps{} differs slightly from~\citep{bateni2020improved}.
Here we describe the key architecture differences which were made with the goal of reducing the number of model parameters.
We verified that these modification came without a reduction in classification performance:
\begin{itemize}[leftmargin=*]
    \setlength\itemsep{0pt}
    \item We replace the ResNet18 \citep{he2016deep} feature extractor with an EfficientNet-B0 \citep{tan2019efficientnet} since it has superior classification performance and fewer parameters (4.0M versus 11.2M for ResNet18). We pre-train the parameters of the feature extractor on ImageNet~\citep{deng2009imagenet} and then freeze them during meta-training and meta-testing.
    \item Like Simple \cnaps{}, we use Feature-wise Linear Modulation (FiLM) layers \citep{perez2018film} to adapt the feature extractor to the current task. In the EfficientNet-B0 feature extractor, we use a FiLM layer with scale parameters $\vgamma_i$ and offset parameters $\vbeta_i$ after every separate convolutional layer and after every depth-wise separable convolution within a inverted residual block (refer to \cref{fig:film}). This is a total of 18 FiLM layers (<0.2\% parameters in the model).
    \item We use a lower capacity 2-layer MLP network for generating parameters for each FiLM layer in the feature extractor (refer to \cref{fig:film_generator}). This new FiLM layer generator network has less than 18\% of the parameters (1.51M versus 8.45M) compared to the network used in the original Simple \cnaps{}.
    \item We do not use the Simple \cnaps{} Auto-regressive (AR) mode as the additional number of parameters did not yield sufficient gain.
\end{itemize}

Since the feature extractor parameters are frozen and the Mahalanobis distance based classifier has no parameters, the only learnable parameters in the model are in the set encoder and the network that generates the FiLM layer parameters.

\section{Experimental Details}
\label{sec:exp_details}
In this section, we provide details for the \lite{ }experiments using the ORBIT and \vtabmd{} datasets.
\subsection{ORBIT Teachable Object Recognition Benchmark}
\label{app:exp_details_for_orbit}
Meta-training and meta-testing for the ORBIT experiments were performed on a single NVIDIA Titan RTX with 24GB of memory.
\paragraph{Feature extractors}
We use either a ResNet-18 (following~\citep{massiceti2021orbit}) or an EfficientNet-B0~\citep{tan2019efficientnet}, both pre-trained on ImageNet~\cite{ILSVRC15}.
Note, for \cnaps{} and Simple \cnaps{}, the feature extractor is frozen and only the set encoder and hyper-networks are trained, for ProtoNets and MAML all parameters are learned, and for the FineTuner the feature extractor is frozen and only the linear classifier is fine-tuned.

\paragraph{Meta-training protocol}
We train the learnable parameters in the meta-learners episodically on \(50\) randomly sampled tasks per train user per epoch (44 total train users).
Note, each epoch samples \(50\) \emph{new} tasks per train user.
Each task is composed of clips sampled from a single user's objects (random way) and associated videos (random shot).
In the case of a large task, following~\citep{massiceti2021orbit}, we randomly sample 4 clips from each support and query video, where each clip is 8 frames.
%
%
For a small task, we limit this to 1 clip per support video and 1 clip per query video where each clip is 8 frames, and we also cap \begin{inparaenum}[1)]
\item the number of objects per task to 5, and 
\item the number of support/query videos per object to 2.
\end{inparaenum}
For both large and small tasks, a clip feature is taken as the average of its frame features, where each frame in $224 \times 224$ pixels.
Note, the FineTuner undergoes no training -- all feature extractor parameters are frozen to its pre-trained weights.

\paragraph{Meta-testing protocol}
Following~\citep{massiceti2021orbit}, we evaluate the trained models on \(5\) tasks per test user, where each task is sampled from just that user's objects and videos.
Different to training, here each task contains \emph{all} the test user's objects and associated videos without caps.
For each test task, we randomly sample 8 clips from each support video, and \emph{all} overlapping clips from each query video.
We then adapt the trained model to a task by using the task's support clips to:
\begin{inparaenum}[i)]
\item perform a forward pass for CNAPs, Simple CNAPs and ProtoNets,
\item take 15 gradient steps on all the model's parameters for MAML, or
\item take 50 gradient steps on just the linear classifier head for FineTuner.
\end{inparaenum}
We evaluate the adapted model predictions for every clip in each query video in the test task (in the clean video evaluation mode, the query videos show just one object on a clear surface, while in the clutter video evaluation mode, the query videos show the object in a multi-object/cluttered scene).
We report all metrics averaged over a flattened list of all the query videos from all tasks from all test users (17 test users, 85 tasks in total), along with its corresponding 95\% confidence interval.

\paragraph{Optimization hyper-parameters}
For CNAPs, Simple CNAPs, and ProtoNets, we use the Adam optimizer~\citep{kingma2014adam} and a learning rate of $10^{\text{-}4}$.
For MAML, we use Adam and a learning rate of \(10^{\text{-}5}\) for the outer loop, and Stochastic Gradient Descent (SGD) and a learning rate of \(10^{\text{-}3}\) for the inner loop (rates reduced by \(0.1\) for the feature extractor in both loops).
For the FineTuner, we use SGD and a learning rate of $0.1$.
We train Simple \cnaps{} with ResNet-18/EfficientNet-B0 for 10/15 epochs respectively, \cnaps{} for 15/15 epochs, ProtoNets for 20/20 epochs, and MAML for 20/20 epochs.
These were chosen based on the number of learnable parameters in each model.
\cref{tab:orbit_results} reports the test performance of the model with the best frame accuracy on a held-out validation set.
\paragraph{\lite{} hyper-parameters}
We train CNAPs, Simple CNAPs and ProtoNets with $H=8$ clips (see~\cref{alg:lite}).
We set the query batch size to $M_b = 8$ clips across all meta-learners.
Note, MAML does not use LITE since we implement only the first-order variant.
We, therefore, process support (and query) sets using standard batch processing with a batch size of $32$ clips. 

\subsection{\vtabmd{} Benchmark}
\label{app:exp_details_for_vtabmd}
Meta-training and meta-testing for the \vtabmd{} experiments were performed on a single NVIDIA V100 16GB GPU. Meta-training takes about 20 hours.
\paragraph{Meta-training protocol}
\scl{} uses an EfficientNet-B0 pretrained on ImageNet for the feature extractor $f$ and all of its parameters are frozen and not updated during meta-training.
As permitted in the \vtabmd{} protocol, we meta-train \scl{} in an episodic manner on the training splits of following datasets: ImageNet, Omniglot, Aircraft, CU Birds, DTD, QuickDraw, and Fungi.
In addition, we meta-train on the test split of MNIST as it does not overlap with any of the test datasets.
We meta-train for 10,000 iterations with the Adam \citep{kingma2014adam} optimizer using a fixed learning rate of 0.001, and a batch size of 40. We back-propagate after every task, but do an optimization step after every 16 tasks.
\paragraph{Meta-testing protocol}
For meta-testing on MD-v2, we generate test episodes using the \metadataset{} episode reader with the standard evaluation settings.
We test all models with 600 episodes each on all test datasets.
The classification accuracy is averaged over the episodes and a 95\% confidence interval is computed.
For each test dataset in VTAB-v2, we use the TensorFlow Datasets API \citep{TFDS} and randomly sample 1000 examples from the train split for the support set and use the entire test split for the query set and report a single accuracy.

\section{Additional Experimental Results}
\label{app:sec:additional_experiments}

\subsection{Full results on ORBIT benchmark}
\label{app:sec:orbit_results_extended}

In the main paper, we report frame and video accuracy for test tasks, as well as the number of MACs and steps to adapt at test time and the number of model parameters.
In~\cref{app:tab:orbit_results_extended}, we include a further metric -- frames to recognition or FTR -- which was proposed in the original baselines~\citep{massiceti2021orbit}.
We also include additional results for large images (224) on small tasks without using \lite{}.
Descriptions for the metrics are thus:
\begin{itemize}[leftmargin=*]
\setlength\itemsep{0pt}
    \item \textbf{Frame accuracy}, the proportion of correct frame predictions in a query video;
    \item \textbf{Frames-to-recognition or FTR}, the number of frames before the first correct prediction, divided by the number of frames in the query video;
    \item \textbf{Video accuracy}, 1 if the most frequent frame prediction in a query video equals the true video label, otherwise 0;
    \item \textbf{MACs to adapt}, number of Multiply-Accumulate operations to learn a new task at test time (i.e.\ the operations required to process the whole support set);
    \item \textbf{Steps to adapt}, number of steps to learn a new task at test time (note, for gradient-based methods this involves multiple forward-backward passes through the model, while for amortization- and metric-based approaches this involves just a single forward pass);
    \item \textbf{Number of parameters}, number of learnable and frozen parameters in the model (note, this exclude the parameters that are generated by amortization-based methods)
\end{itemize}
\input{tables/orbit_results_extended}

\subsection{Tabular results on \vtabmd{} benchmark}
\label{app:sec:vtab_tabular_results}

In~\cref{tab:vtab+md_details}, we show the tabular results for the \vtabmd{} benchmark.
%
\input{tables/vtab_md_results_table}
\subsection{Meta-training without LITE on small tasks with large images}
\label{app:sec:small_tasks}
\input{tables/vtab+md_ablation_table}

\cref{tab:vtab+md_ablation} we show classification results on \vtabmd{} using various ablations of Simple \cnaps{} including \lite{} on versus off, image size 84$\times$84 versus 256$\times$256 pixels, and small versus large sized tasks. For the no \lite{}, image size 84$\times$84, and large task case, we meta-train on 35,000 tasks using the Adam optimizer at learning rate of 0.001 on the same training datasets as \scl{}. For the no \lite{}, image size 224$\times$224 pixels, and small task case, we we meta-train on 15,000 tasks using the Adam optimizer at learning rate of 0.001 on the same training datasets as \scl{}. To make the number of tasks small during meta-training, we limit the maximum support set size  to be 40 and the maximum classification way to be 30.

It is clear that using larger images results in a significant boost in classification accuracy, except on datasets where the images are natively small (e.g.~Omniglot, Quickdraw, dSprites). Using \lite{} versus a smaller task size results in a significant boost in classification accuracy on VTAB-v2 where the support set size is large (1000 examples), however the results on MD-v2, where the support set sizes are smaller, are very similar in the two cases.

The trend is similar in the case of the ORBIT dataset (refer to \cref{app:tab:orbit_results_extended}) where the difference between using \lite{} and tasks with a smaller number of examples is not great (often within the margin of error). This is likely due to the fact that in the case of ORBIT
\begin{inlinelist}
\item the classification way is typically small (less than or equal to 10); and
\item since the support frames are derived from videos, there is significant redundancy in the support sets, making the difference between having a small and large number of support examples less important.
\end{inlinelist} 
The benefits of \lite{} are more apparent in tasks with large way and large support set sizes, as is the case with VTAB-v2.
\subsection{Tabular results and additional details on the varying $|\mathcal{H}|$ experiments}
\label{app:varying_h}
\cref{tab:vtab+md_vary_h,tab:vtab+md_vary_h_protonets,tab:vtab+md_vary_h_84} provide full tabular results for classification accuracy versus varying $|\mathcal{H}|$ on \vtabmd{}.
Note that in \cref{tab:vtab+md_vary_h_84}, using \scl{} on images of size of $84 \times 84$ pixels with $|\mathcal{H}|=40$, GPU memory usage drops to roughly 8 GB, which is approximately half of that used when run without \lite{} (i.e. $|\mathcal{H}|=|\mathcal{D}_S|$).

\cref{tab:mse_vs_h} shows the mean squared error between the mean of the approximate gradients and the true gradients for both \lite{} and sub-sampled small tasks as $|\mathcal{H}|$ is varied.
%
%
The low mean squared error values for both training methods empirically demonstrates that both are unbiased.

\cref{tab:std_vs_h} shows the average root mean squared error (RSME) of the approximate gradients and the true gradients for both \lite{} and sub-sampled small tasks as $|\mathcal{H}|$ is varied.
\cref{tab:std_vs_h} and \cref{fig:std_vs_h} show that the RMSE deviation of the \lite{} estimate is significantly smaller than that of sub-sampled small tasks at all but the highest values of $|\mathcal{H}|$.
Note, that these results are limited to image classification in the specific networks and network parameters that we tested. Other data types and networks are left for future work.

These experiments were carried out as follows:
\begin{itemize}
    \item The \scl{} network is initialized identically for all runs.
    \item Image size is $84 \times 84$ pixels, so that the true gradients can be calculated.
    \item The same 10-way, 10-shot task ($|\mathcal{D}_S|=100$) drawn from the DTD dataset is identical for all runs.
    \item Gradients are measured on the weights in the first (i.e. earliest) Conv2D layer in the set encoder after a single training iteration.
    \item Reference (exact) gradients are calculated without using LITE.
    \item Small task gradients are calculated by randomly sub-sampling the task (though we ensure there is at least one example per class).
    \item For each value of $|\mathcal{H}|$, the number of samples used in the calculations are chosen such that 1000 examples of the support set are used. For example, for $|\mathcal{H}|=50$, 20 different one iteration training runs are done (20 runs $\times$ 50 random examples per run = 1000).
    \item To calculate the values in \cref{tab:mse_vs_h}, for each value of $|\mathcal{H}|$, the mean of the approximate gradient runs is computed and then the mean squared error is computed between this value and the exact gradient.
    \item To calculate the values in \cref{tab:std_vs_h}, for each value of $|\mathcal{H}|$, the RMSE between the approximate gradients and the exact gradients is computed and then this value is averaged over the number of runs.
\end{itemize}
\input{tables/vtab+md_vary_H}
%
\input{tables/vtab+md_vary_H_protonets}
%
\input{tables/vtab+md_vary_H_84}
%
\input{tables/mean_std_vs_h}
\subsection{Additional Results}
\label{app:additional_results}
\cref{tab:vtab+md_vary_330} contains the results for \scl{} on \vtabmd{} at image size $320 \times 320$ pixel with $|\mathcal{H}|=10$, demonstrating that by employing \lite{}, even larger images can be be used in meta-learning algorithms. Overall, these results are similar to the $224 \times 224$ case as the feature extractor was pre-trained at $224 \times 224$ pixels. However, on the Birds, Fungi, and Retinopathy datasets, where the original images are very large ($> 320$ pixels), the results on this run were better than the $224$ case.
%
\input{tables/vtab+md_330}
\end{document}

%% file: math_commands.tex
\usepackage{amsmath,amsfonts,bm}

\def\eqref#1{equation~\ref{#1}}

\def\1{\bm{1}}

\def\vtheta{{\bm{\theta}}}

\def\vphi{{\bm{\phi}}}

\def\vgamma{{\bm{\gamma}}}
\def\vbeta{{\bm{\beta}}}

\def\vx{{\bm{x}}}

\DeclareMathAlphabet{\mathsfit}{\encodingdefault}{\sfdefault}{m}{sl}
\SetMathAlphabet{\mathsfit}{bold}{\encodingdefault}{\sfdefault}{bx}{n}



%% file: tables/orbit_results.tex
\begin{table}[htbp]
    \vspace{-1em}
    \setlength\tabcolsep{1.2pt}
    \caption{\textbf{Training meta-learners on large images with \lite{} achieves state-of-the-art accuracy with low test time adaption cost on  ORBIT.} Results are reported as the average (95\% confidence interval) over 85 test tasks (5 tasks per test user, 17 test users). $I$ is image size. $f$ is model trained with/without \lite{}. RN-18 is ResNet-18. EN-B0 is EfficientNet-B0. T is \(\times\!\)10\textsuperscript{12} MACs. F is forward pass. FB is forward-backward pass. Time is average wall clock time per task in seconds.
    }
    \vspace{0.9em}
    \label{tab:orbit_results}
    \centering
    \scalebox{0.7}{
    \begin{tabular}{lclP{1.9cm}P{1.8cm}P{0.3cm}P{1.9cm}P{1.8cm}P{0.3cm}P{1.4cm}P{1.4cm}P{1.4cm}P{1.2cm}}
    \toprule
    & & & \multicolumn{2}{c}{\textbf{Clean Videos}} && \multicolumn{2}{c}{ \textbf{Clutter Videos}} && \multicolumn{3}{c}{\textbf{Test-time adaption}}\\
    \cmidrule{4-5}\cmidrule{7-8}\cmidrule{10-12}
    \textsc{\textbf{model}} & $I$ & $f$  & \textsc{\textbf{frame acc}} $\uparrow$ & \textsc{\textbf{video acc}} $\uparrow$ && \textsc{\textbf{frame acc}} $\uparrow$ & \textsc{\textbf{video acc}} $\uparrow$ && {\textbf{\textsc{macs}} \quad\qquad $\downarrow$} & \textsc{\textbf{steps} \quad\qquad $\downarrow$} & \textsc{\textbf{time} \quad\qquad $\downarrow$} & \textsc{\textbf{params} $\downarrow$} \\
    \cmidrule{1-13}
    \multirow{3}{*}{FineTuner~\citep{yosinski2014transferable}} & 84 & RN-18 & 69.5 (2.2) & 79.7 (2.6) && 53.7 (1.8) & 63.1 (2.4) && 317.70T & 50FB & 53.94s & 11.17M \\
    & 224 & RN-18 & 72.2 (2.2) & 81.9 (2.6) && 56.7 (2.0) & 61.3 (2.5) && 546.57T & 50FB & 96.23s & 11.18M  \\
    & 224 & EN-B0 & \textbf{78.1 (2.0)} & 85.9 (2.3) && \textbf{63.1 (1.8)} & 66.9 (2.4) && 121.02T & 50FB & 139.99s & \textbf{4.01M}  \\
    \cmidrule{1-13}
    \multirow{3}{*}{MAML~\citep{finn2017model}} & 84 & RN-18 & 70.6 (2.1) & 80.9 (2.6) &&  51.7 (1.9) & 57.9 (2.5) && 95.31T & 15FB & 36.98s & 11.17M   \\
    & 224 & RN-18 & 75.7 (1.9) & 86.1 (2.3) && 59.3 (1.9) & 64.3 (2.4) && 163.97T & 15FB & 65.22s & 11.18M    \\
    & 224 & EN-B0 & \textbf{79.3 (1.9)} & 87.5 (2.2) && \textbf{64.6 (1.9)} & 69.4 (2.3) && 36.31T & 15FB & 117.89s & \textbf{4.01M} \\
    \cmidrule{1-13}
    \multirow{3}{*}{ProtoNets~\citep{snell2017prototypical}} & 84 & RN-18 & 65.2 (2.0) & 81.9 (2.5) && 50.3 (1.7) & 59.9 (2.5) && 3.18T & \textbf{1F} & \textbf{0.73s} & 11.17M \\
    & 224 & RN-18 \kern.11em+ \lite{} & 76.7 (1.9) & 86.4 (2.2) && 61.4 (1.8) & \textbf{68.5 (2.4)} && 5.47T  & \textbf{1F} & 1.07s & 11.18M \\
    & 224 & EN-B0 + \lite{} & \textbf{82.1 (1.7)} & \textbf{91.2 (1.9)} && \textbf{66.3 (1.8)} & \textbf{72.9 (2.3)} && \textbf{1.21T} & \textbf{1F} & 1.72s & \textbf{4.01M}  \\
    \cmidrule{1-13}
    \multirow{3}{*}{CNAPs~\citep{requeima2019cnaps}} & 84 & RN-18 & 66.2 (2.1) & 79.6 (2.6) && 51.5 (1.8) & 59.5 (2.5) && 3.48T & \textbf{1F} & 0.98s & 12.75M \\
    & 224 & RN-18 \kern.11em+ \lite{} & 76.0 (1.9) & 84.9 (2.3) && 58.2 (1.9) & 62.5 (2.5) && 7.64T & \textbf{1F} & 2.11s & 12.76M  \\
    & 224 & EN-B0 + \lite{} & \textbf{79.6 (1.9)} & 87.6 (2.2) && \textbf{63.3 (1.9)} & \textbf{69.2 (2.3)} && 3.38T & \textbf{1F} & 2.85s & 10.59M \\
    \cmidrule{1-13}
    \multirow{3}{*}{Simple CNAPs~\citep{bateni2020improved}} & 84 & RN-18 & 70.3 (2.1) & 83.0 (2.5) && 53.9 (1.8) & 62.0 (2.5) && 3.48T & \textbf{1F} & 1.01s & 11.97M  \\
    & 224 & RN-18 \kern.11em+ \lite{} & 76.5 (2.0) & 86.4 (2.2) && 57.5 (1.9) & 64.6 (2.4) && 7.64T & \textbf{1F} & 2.14s & 11.97M \\
    & 224 & EN-B0 + \lite{} & \textbf{82.7 (1.7)} & \textbf{91.8 (1.8)} && \textbf{65.6 (1.9)} & \textbf{71.9 (2.3)} && 3.39T & \textbf{1F} & 2.92s & 5.67M \\
    \bottomrule
    \end{tabular}}
    \vspace{0.5em}
\end{table}

%% file: tables/varying_h_summary.tex
\begin{table}[htbp]
  \centering
  \caption{Classification accuracy results in percent on \vtabmd{} using Simple CNAPs and ProtoNets with varying values of $|\mathcal{H}|$. Image sizes are $224 \times 224$ and $84 \times 84$ pixels. For $|\mathcal{H}| > 40$, we used gradient/activation checkpointing methods \citep{chen2016training} in addition to LITE. For full results see \cref{app:varying_h}}
    \begin{adjustbox}{max width=\textwidth}
    \begin{tabular}{lcccccc|ccccc|cc}
    \toprule
    Model      & \multicolumn{6}{c}{Simple CNAPs}              & \multicolumn{5}{c}{ProtoNets}         & \multicolumn{2}{c}{Simple CNAPs} \\
    Image Size & \multicolumn{6}{c}{$224 \times 224$}                 & \multicolumn{5}{c}{$224 \times 224$}         & \multicolumn{2}{c}{$84 \times 84$} \\
    \midrule
    $|\mathcal{H}|$     & 1     & 10    & 20    & 30    & 40    & 100   & 0     & 10    & 20    & 30    & 40    & \multicolumn{1}{c}{40} & $|\mathcal{D}_S|$ \\
    \midrule
    MD-v2 & 72.8  & 73.7  & 73.3  & 73.8  & 73.9  & 74.3  & 71.0    & 72.0    & 72.0    & 72.5  & 72.7  & 63.6  & 68.4 \\
    VTAB (all) & 51.2  & 51.0    & 50.5  & 51.1  & 51.4  & 51.2  & 45.1  & 45.8  & 46.2  & 46.2  & 46.1  & 42.7  & 44.7 \\
    VTAB (natural) & 64.5  & 65.3  & 64.1  & 65.8  & 65.2  & 66.0    & 58.5  & 60.0    & 60.6  & 60.8  & 60.9  & 47.7  & 49.5 \\
    VTAB (specialized) & 71.8  & 71.4  & 70.5  & 71.3  & 71.9  & 71.6  & 63.5  & 63.9  & 64.2  & 64.5  & 64.2  & 61.0    & 63.8 \\
    VTAB (structured) & 31.0    & 30.0    & 30.3  & 29.9  & 30.8  & 29.9  & 26.0    & 26.2  & 26.4  & 26.1  & 25.9  & 29.9  & 31.7 \\
    \bottomrule
    \end{tabular}%
    \end{adjustbox}
  \label{tab:vary_h_summary}%
\end{table}%

%% file: tables/orbit_results_extended.tex
\begin{table}[htbp]
    \vspace{-1em}
    \setlength\tabcolsep{1.2pt}
    \caption{Training meta-learners on large images with \lite{} achieves state-of-the-art accuracy with low test time adaption cost on the ORBIT Teachable Object Recognition Benchmark. Results are reported as the average (95\% confidence interval) over 85 test tasks (5 tasks per test user, 17 test users). $I$ is image size. $f$ is model trained with/without \lite{}. RN-18 is ResNet-18. EN-B0 is EfficientNet-B0. T is \(\times\!\)10\textsuperscript{12} MACs. F is forward pass. FB is forward-backward pass. Time is average wall clock time per task in seconds.
    }
    \vspace{0.9em}
    \label{app:tab:orbit_results_extended}
    \centering
    \scalebox{0.6}{
    \begin{tabular}{lclP{1.9cm}P{1.4cm}P{1.8cm}P{0.3cm}P{1.9cm}P{1.4cm}P{1.8cm}P{0.3cm}P{1.4cm}P{1.4cm}P{1.4cm}P{1.2cm}}
    \toprule
    & & & \multicolumn{3}{c}{\textbf{Clean Videos}} && \multicolumn{3}{c}{ \textbf{Clutter Videos}} && \multicolumn{3}{c}{\textbf{Test-time adaption}}\\
    \cmidrule{4-6}\cmidrule{8-10}\cmidrule{12-14}
    \textsc{\textbf{model}} & $I$ & $f$  & \textsc{\textbf{frame acc}} $\uparrow$ & \textsc{\textbf{ftr}}\qquad\quad $\downarrow$ & \textsc{\textbf{video acc}} $\uparrow$ && \textsc{\textbf{frame acc}} $\uparrow$ & \textsc{\textbf{ftr}}\quad\qquad $\downarrow$ & \textsc{\textbf{video acc}} $\uparrow$ && {\textbf{\textsc{macs}}}\quad\qquad $\downarrow$ & \textsc{\textbf{steps}}\quad\qquad $\downarrow$ & \textsc{\textbf{time}}\quad\qquad $\downarrow$ & \textsc{\textbf{params}} $\downarrow$ \\
    \cmidrule{1-15}
    \multirow{3}{*}{FineTuner~\citep{yosinski2014transferable}} & 84 & RN-18 & 69.5 (2.2) & 7.8 (1.5) & 79.7 (2.6) && 53.7 (1.8) & 14.4 (1.5) & 63.1 (2.4) && 317.70T & 50FB & 53.94s & 11.17M \\
    & 224 & RN-18 & 72.2 (2.2) & 8.7 (17) & 81.9 (2.6) && 56.7 (2.0) & 18.8 (1.8) & 61.3 (2.5) && 546.57T & 50FB & 96.23s & 11.18M  \\
    & 224 & EN-B0 & \textbf{78.1 (2.0)} & \textbf{5.8 (1.4)} & 85.9 (2.3) && \textbf{63.1 (1.8)} & \textbf{11.5 (1.4)} & 66.9 (2.4) && 121.02T & 50FB & 139.99s & \textbf{4.01M}  \\
    \cmidrule{1-15}
    \multirow{3}{*}{MAML~\citep{finn2017model}} & 84 & RN-18 & 70.6 (2.1) & 8.6 (1.6) & 80.9 (2.6) &&  51.7 (1.9) & 21.0 (1.8) & 57.9 (2.5) && 95.31T & 15FB & 36.98s & 11.17M   \\
    & 224 & RN-18 & 75.7 (1.9) & \textbf{4.9 (1.2)} & 86.1 (2.3) && 59.3 (1.9) & 16.3 (1.7) & 64.3 (2.4) && 163.97T & 15FB & 65.22s & 11.18M    \\
    & 224 & EN-B0 & \textbf{79.3 (1.9)} & \textbf{6.2 (1.4)} & 87.5 (2.2) && \textbf{64.6 (1.9)} & \textbf{12.8 (1.5)} & 69.4 (2.3) && 36.31T & 15FB & 117.89s & \textbf{4.01M} \\
    \cmidrule{1-15}
    \multirow{5}{*}{ProtoNets~\citep{snell2017prototypical}} & 84 & RN-18 & 65.2 (2.0) & 7.6 (1.4) & 81.9 (2.5) && 50.3 (1.7) & 14.9 (1.5) & 59.9 (2.5) && 3.18T & \textbf{1F} & \textbf{0.73s} & 11.17M \\
    & 224 & RN-18 & 77.4 (1.8) & \textbf{4.5 (1.1)} & 87.1 (2.2) && 56.8 (1.8) & 14.4 (1.5) & 62.5 (2.5) && 5.47T & \textbf{1F} & 1.07s & 11.18M \\
    & 224 & RN-18 \kern.11em+ \lite{} & 76.7 (1.9) & \textbf{5.1 (1.2)} & 86.4 (2.2) && 61.4 (1.8) & \textbf{13.2 (1.5)} & \textbf{68.5 (2.4)} && 5.47T  & \textbf{1F} & 1.07s & 11.18M \\
    & 224 & EN-B0 & \textbf{78.4 (1.8)} & \textbf{4.7 (1.1)} & \textbf{87.9 (2.1)} && 57.3 (1.8) & \textbf{12.7 (1.4)} & 63.9 (2.4) && \textbf{1.21T} & \textbf{1F} & 1.72s & \textbf{4.01M}\\
    & 224 & EN-B0 + \lite{} & \textbf{82.1 (1.7)} & \textbf{3.9 (1.0)} & \textbf{91.2 (1.9)} && \textbf{66.3 (1.8)} & \textbf{12.7 (1.5)} & \textbf{72.9 (2.3)} && \textbf{1.21T} & \textbf{1F} & 1.72s & \textbf{4.01M}  \\
    \cmidrule{1-15}
    \multirow{5}{*}{CNAPs~\citep{requeima2019cnaps}} & 84 & RN-18 & 66.2 (2.1) & 8.4 (1.4) & 79.6 (2.6) && 51.5 (1.8) & 17.9 (1.7) & 59.5 (2.5) && 3.48T & \textbf{1F} & 0.98s & 12.75M \\
    & 224 & RN-18 & 73.6 (2.0) & \textbf{5.4 (1.2)} & 83.4 (2.4) && 57.6 (1.8) & 14.9 (1.6) & 66.5 (2.4) && 7.64T & \textbf{1F} & 2.11s & 12.76M \\
    & 224 & RN-18 \kern.11em+ \lite{} & 76.0 (1.9) & \textbf{5.9 (1.3)} & 84.9 (2.3) && 58.2 (1.9) & 15.1 (1.6) & 62.5 (2.5) && 7.64T & \textbf{1F} & 2.11s & 12.76M  \\
    & 224 & EN-B0 & \textbf{79.6 (1.9)} & \textbf{6.2 (1.4)} & 87.0 (2.2) && \textbf{62.6 (1.9)} & \textbf{13.2 (1.5)} & 67.4 (2.4) && 3.38T & \textbf{1F} & 2.83s & 10.59M \\
    & 224 & EN-B0 + \lite{} & \textbf{79.6 (1.9)} & \textbf{5.9 (1.3)} & 87.6 (2.2) && \textbf{63.3 (1.9)} & \textbf{12.8 (1.5)} & \textbf{69.2 (2.3)} && 3.38T & \textbf{1F} & 2.85s & 10.59M \\
    \cmidrule{1-15}
    \multirow{5}{*}{Simple CNAPs~\citep{bateni2020improved}} & 84 & RN-18 & 70.3 (2.1) & 7.3 (1.5) & 83.0 (2.5) && 53.9 (1.8) & 16.0 (1.6) & 62.0 (2.5) && 3.48T & \textbf{1F} & 1.01s & 11.97M  \\
    & 224 & RN-18 & 75.2 (2.0) & \textbf{6.0 (1.4)} & 84.6 (2.4) && 58.1 (1.9) & 14.7 (1.6) & 60.9 (2.5) && 7.64T & \textbf{1F} & 2.13s & 11.97M  \\
    & 224 & RN-18 \kern.11em+ \lite{} & 76.5 (2.0) & \textbf{6.1 (1.4)} & 86.4 (2.2) && 57.5 (1.9) & 17.3 (1.7) & 64.6 (2.4) && 7.64T & \textbf{1F} & 2.14s & 11.97M \\
    & 224 & EN-B0 & \textbf{81.4 (1.8)} & \textbf{4.9 (1.3)} & \textbf{88.3 (2.1)} && \textbf{65.6 (1.9)} & \textbf{11.2 (1.4)} & \textbf{69.9 (2.3)} && 3.39T & \textbf{1F} & 2.91s & 5.67M  \\
    & 224 & EN-B0 + \lite{} & \textbf{82.7 (1.7)} & \textbf{4.1 (1.1)} & \textbf{91.8 (1.8)} && \textbf{65.6 (1.9)} & \textbf{13.5 (1.5)} & \textbf{71.9 (2.3)} && 3.39T & \textbf{1F} & 2.92s & 5.67M \\
    \bottomrule
    \end{tabular}}
    \vspace{0.5em}
\end{table}

%% file: tables/vtab_md_results_table.tex
    
\begin{table}[htbp]
  \centering
  \caption{Classification accuracy results on \vtabmd{} \citep{dumoulin2021comparing} using \scl{} and various competing transfer learning and meta-learning approaches. All competitive results are from \citep{dumoulin2021comparing}. All figures are percentages and the $\pm$ sign indicates the 95\% confidence interval over tasks. Bold type indicates the highest scores (within the confidence interval). The \textsc{VTAB-v2} results have no confidence interval as the testing protocol requires only a single run over the entire test set. RN indicates ResNet \citep{he2016deep} and EN indicates EfficientNet \citep{tan2019efficientnet}. The SUR results are with a linear classifier head. \scl{} outperforms all approaches on MD-v2 and outperforms all meta-learning approaches on VTAB (all).}
    \vspace{0.9em}
    \begin{adjustbox}{max width=\textwidth}
    \begin{tabular}{lcccccccc}
    \toprule
          & \multicolumn{3}{c}{Transfer learning} & \multicolumn{5}{c}{Meta-Learning} \\
          &  MD-Transfer &  SUR  &  BiT  &  ProtoNets &  ProtoMAML &  CTX  & SC(84) & SC+LITE \\
    \midrule
    Backbone & RN-18 & RN-50 x 7 & RN-18 & RN-18 & RN-18 & RN-34 & EN-B0 & EN-B0 \\
    Params (M) & 11.2M & 164.6M & 11.2M & 11.2M & 11.2M & 21.3M & 4.0M  & 4.0M \\
    Image Size & 126   & 224   & 224   & 126   & 126   & 224   & 84    & 224 \\
    \midrule
    Omniglot &  82.0 $\pm$ 1.3 & 89.6  &  72.7 $\pm$ 4.6 &  85.3 $\pm$ 0.9 &  \textbf{90.2 $\pm$ 0.7} &  84.6 $\pm$ 0.9 & \textbf{90.9 $\pm$ 0.6} & 86.5 $\pm$ 0.8 \\
    Aircraft &  76.8 $\pm$ 1.2 & 59.7  &  73.6 $\pm$ 3.8 &  74.3 $\pm$ 0.8 &  82.1 $\pm$ 0.6 &  \textbf{85.3 $\pm$ 0.8} & 77.5 $\pm$ 0.7 & 83.6 $\pm$ 0.7 \\
    Birds &  61.2 $\pm$ 1.3 & 81.4  &  \textbf{87.2 $\pm$ 1.9} &  68.0 $\pm$ 1.0 &  73.4 $\pm$ 0.9 &  72.9 $\pm$ 1.1 & 76.4 $\pm$ 0.8 & \textbf{88.6 $\pm$ 0.7} \\
    DTD   &  66.0 $\pm$ 1.1 & \textbf{83.9}  &  \textbf{82.6 $\pm$ 2.7} &  65.3 $\pm$ 0.7 &  66.3 $\pm$ 0.8 &  77.3 $\pm$ 0.7 & 74.3 $\pm$ 0.7 & \textbf{84.1 $\pm$ 0.7} \\
    QuickDraw &  61.3 $\pm$ 1.1 & \textbf{81.2}  &  66.3 $\pm$ 3.6 &  60.6 $\pm$ 1.0 &  66.4 $\pm$ 1.0 &  73.3 $\pm$ 0.8 & 76.5 $\pm$ 0.7 & 75.7 $\pm$ 0.8 \\
    Fungi &  35.5 $\pm$ 1.1 & \textbf{69.2}  &  53.9 $\pm$ 4.4 &  39.8 $\pm$ 1.1 &  46.3 $\pm$ 1.1 &  48.0 $\pm$ 1.2 & 51.3 $\pm$ 1.1 & 56.9 $\pm$ 1.2 \\
    Traffic Sign &  \textbf{84.7 $\pm$ 0.9} & 46.5  &  75.4 $\pm$ 4.3 &  49.8 $\pm$ 1.1 &  50.3 $\pm$ 1.1 &  80.1 $\pm$ 1.0 & 54.8 $\pm$ 1.1 & 65.8 $\pm$ 1.1 \\
    MSCOCO &  39.6 $\pm$ 1.0 & \textbf{58.6}  &  \textbf{60.0 $\pm$ 2.9} &  39.7 $\pm$ 1.0 &  39.0 $\pm$ 1.0 &  51.4 $\pm$ 1.1 & 45.1 $\pm$ 1.0 & 50.0 $\pm$ 1.0 \\
    \midrule
    Caltech101 & 70.6  & 86.5  & 84.6  & 72.0    & 73.1  & 84.2  & 79.6  & \textbf{87.7} \\
    CIFAR100 & 31.3  & 34.2  & 47.1  & 27.7  & 29.7  & 37.5  & 37.1  & \textbf{48.8} \\
    Flowers102 & 66.1  & 71.2  & 82.7  & 57.1  & 60.2  & 81.8  & 65.5  & \textbf{83.5} \\
    Pets  & 49.1  & 88.7  & 83.9  & 51.0    & 56.6  & 70.9  & 69.8  & \textbf{89.3} \\
    Sun397 & 13.9  & 0.5   & 29.1  & 14.2  & 8.1   & 24.8  & 18.0  & \textbf{30.9} \\
    SVHN  & 83.2  & 24.2  & \textbf{83.4}  & 41.9  & 46.8  & 67.2  & 26.7  & 51.0 \\
    \midrule
    EuroSAT & 88.7  & 82.6  & \textbf{93.8}  & 77.7  & 80.1  & 86.4  & 82.8  & 89.3 \\
    Resics45 & 63.7  & 67.8  & 74.1  & 50.8  & 53.5  & 67.7  & 64.5  & \textbf{76.4} \\
    Patch Camelyon & \textbf{81.5}  & 77.1  & 80.7  & 73.8  & 75.9  & 79.8  & 78.4  & 81.4 \\
    Retinopathy & 57.6  & 37.4  & \textbf{74.5}  & 28.0    & 73.2  & 35.5  & 29.4  & 40.3 \\
    \midrule
    CLEVR-count & 40.3  & 34.1  & \textbf{55.2}  & 32.0    & 32.7  & 27.9  & 30.7  & 31.4 \\
    CLEVR-dist & 52.9  & 29.8  & \textbf{58.7}  & 39.4  & 35.4  & 29.6  & 32.5  & 32.8 \\
    dSprites-loc & 85.9  & 16.9  & \textbf{98.6}  & 38.1  & 42.0 & 23.2  & 43.9  & 12.3 \\
    dSprites-ori & 46.4  & 18.7  & 46.5  & 16.3  & 23.0 & \textbf{46.9}  & 21.1  & 31.1 \\
    SmallNORB-azi & 36.5  & 8.3   & 20.1  & 12.3  & 13.4  & \textbf{37.0} & 13.5  & 14.5 \\
    SmallNORB-elev & \textbf{31.2}  & 18.4  & 21.8  & 17.4  & 18.8  & 21.6  & 19.6  & 21.0 \\
    DMLab & 43.0  & 33.5  & \textbf{43.7}  & 31.8  & 32.5  & 31.9  & 33.9  & 39.4 \\
    KITTI-dist & 58.7  & 57.5  & \textbf{78.8}  & 42.1  & 54.4  & 54.3  & 58.1  & 63.9 \\
    \midrule
    MD-v2 & 63.4  & 71.3  & 71.5  & 60.3  & 64.2  & 71.6  & 68.4  & \textbf{73.9} \\
    VTAB (all) & 55.6  & 43.7  & \textbf{64.3}  & 40.2  & 45.0  & 50.5  & 44.7  & 51.4 \\
    VTAB (natural) & 52.4  & 50.9  & \textbf{68.5}  & 44.0  & 45.7  & 61.1  & 49.5  & 65.2 \\
    VTAB (specialized) & 72.9  & 66.2  & \textbf{80.8}  & 57.6  & 70.7  & 67.3  & 63.8  & 71.9 \\
    VTAB (structured) & 49.4  & 27.2  & \textbf{53.0}  & 28.7  & 31.5  & 34.1  & 31.7  & 30.8 \\
    \bottomrule
    \end{tabular}%
    \end{adjustbox}
  \label{tab:vtab+md_details}%
\end{table}%

%% file: tables/vtab+md_ablation_table.tex
\begin{table}[htbp]
  \centering
  \caption{Classification accuracy results on \vtabmd{} \citep{dumoulin2021comparing} using various ablations of Simple \cnaps{} including \lite{} on versus off, image size 84$\times$84 versus 256$\times$256 pixels, and small versus large tasks. All figures are percentages and the $\pm$ sign indicates the 95\% confidence interval over tasks. Bold type indicates the highest scores (within the confidence interval). The \textsc{VTAB-v2} results have no confidence interval as the testing protocol requires only a single run over the entire test set. A pretrained EfficientNet-B0 \citep{tan2019efficientnet} backbone was utilized in all runs. In general, using larger images leads to better results, and using \lite{} on large tasks greatly improves results on \textsc{VTAB-v2}.}
    \vspace{0.9em}
    \begin{adjustbox}{max width=0.7\textwidth}
    \begin{tabular}{lccc}
    \toprule
    \lite{} & No & No & Yes \\
    Task Size & Large & Small & Large \\
    Image Size & 84    & 224   & 224 \\
    \midrule
    Omniglot & \textbf{90.9 $\pm$ 0.6} & \textbf{91.6 $\pm$ 0.6} & 86.5 $\pm$ 0.8 \\
    Aircraft & 77.5 $\pm$ 0.7 & 81.5 $\pm$ 0.7 & \textbf{83.6 $\pm$ 0.7} \\
    Birds & 76.4 $\pm$ 0.8 & \textbf{88.8 $\pm$ 0.6} & \textbf{88.6 $\pm$ 0.7} \\
    DTD   & 74.3 $\pm$ 0.7 & \textbf{83.7 $\pm$ 0.6} & \textbf{84.1 $\pm$ 0.7} \\
    QuickDraw & \textbf{76.5 $\pm$ 0.7} & \textbf{76.4 $\pm$ 0.7} & \textbf{75.7 $\pm$ 0.8} \\
    Fungi & 51.3 $\pm$ 1.1 & \textbf{59.3 $\pm$ 1.1} & 56.9 $\pm$ 1.2 \\
    Traffic Sign & 54.8 $\pm$ 1.1 & 60.7 $\pm$ 1.0 & \textbf{65.8 $\pm$ 1.1} \\
    MSCOCO & 45.1 $\pm$ 1.0 & \textbf{52.5 $\pm$ 1.1} & 50.0 $\pm$ 1.0 \\
    \midrule
    Caltech101 & 79.6  & 84.9  & \textbf{87.7} \\
    CIFAR100 & 37.1  & \textbf{50.2}  & 48.8 \\
    Flowers102 & 65.5  & 78.9  & \textbf{83.5} \\
    Pets  & 69.8  & 87.7  & \textbf{89.3} \\
    Sun397 & 18.0    & \textbf{32.0}    & 30.9 \\
    SVHN  & 26.7  & 37.6  & \textbf{51.0} \\
    \midrule
    EuroSAT & 82.8  & 86.0    & \textbf{89.3} \\
    Resics45 & 64.5  & 69.8  & \textbf{76.4} \\
    Patch Camelyon & 78.4  & 79.1  & \textbf{81.4} \\
    Retinopathy & 29.4  & 40.2  & \textbf{40.3} \\
    \midrule
    CLEVR-count & 30.7  & 28.7  & \textbf{31.4} \\
    CLEVR-dist & 32.5  & 31.4  & \textbf{32.8} \\
    dSprites-loc & \textbf{43.9}  & 14.7  & 12.3 \\
    dSprites-ori & 21.1  & \textbf{35.8}  & 31.1 \\
    SmallNORB-azi & 13.5  & 12.2  & \textbf{14.5} \\
    SmallNORB-elev & 19.6  & 19.0    & \textbf{21.0} \\
    DMLab & 33.9  & 36.7  & \textbf{39.4} \\
    KITTI-dist & 58.1  & 57.0    & \textbf{63.9} \\
    \midrule
    MD-v2 & 68.4  & \textbf{74.3}  & 73.9 \\
    VTAB (all) & 44.7  & 49.0    & \textbf{51.4} \\
    VTAB (natural) & 49.5  & 61.9  & \textbf{65.2} \\
    VTAB (specialized) & 63.8  & 68.8  & \textbf{71.9} \\
    VTAB (structured) & \textbf{31.7}  & 29.4  & 30.8 \\
    \bottomrule
    \end{tabular}%
    \end{adjustbox}
  \label{tab:vtab+md_ablation}%
\end{table}%

%% file: tables/vtab+md_vary_H.tex
\begin{table}[htbp]
  \centering
  \caption{Classification accuracy results on \vtabmd{} \citep{dumoulin2021comparing} using \scl{} with varying values of $|\mathcal{H}|$. Image size is 224 x 224 pixels. To achieve $|\mathcal{H}| > 40$, we used gradient/activation checkpointing methods \citep{chen2016training} in addition to LITE. All figures are percentages and the $\pm$ sign indicates the 95\% confidence interval over tasks. The \textsc{VTAB-v2} results have no confidence interval as the testing protocol requires only a single run over the entire test set.}
    \vspace{0.9em}
    \begin{adjustbox}{max width=\textwidth}
    \begin{tabular}{lcccccc}
    \toprule
    Dataset      & $|\mathcal{H}|=1$ & $|\mathcal{H}|=10$ & $|\mathcal{H}|=20$ & $|\mathcal{H}|=30$ & $|\mathcal{H}|=40$ & $|\mathcal{H}|=100$ \\
    \midrule
    Omniglot & 83.5 $\pm$ 1.0 & 85.2 $\pm$ 0.9 & 85.5 $\pm$ 0.9 & 85.9 $\pm$ 0.9 & 86.5 $\pm$ 0.8 & 86.2 $\pm$ 0.8 \\
    Aircraft & 82.1 $\pm$ 0.8 & 82.9 $\pm$ 0.8 & 82.5 $\pm$ 0.8 & 83.5 $\pm$ 0.7 & 83.6 $\pm$ 0.7 & 83.4 $\pm$ 0.8 \\
    Birds & 88.0 $\pm$ 0.7 & 89.4 $\pm$ 0.5 & 88.9 $\pm$ 0.6 & 88.5 $\pm$ 0.7 & 88.6 $\pm$ 0.7 & 88.8 $\pm$ 0.7 \\
    DTD   & 84.4 $\pm$ 0.7 & 84.3 $\pm$ 0.7 & 84.2 $\pm$ 0.7 & 85.1 $\pm$ 0.6 & 84.1 $\pm$ 0.7 & 85.1 $\pm$ 0.7 \\
    QuickDraw & 75.3 $\pm$ 0.8 & 75.8 $\pm$ 0.8 & 75.7 $\pm$ 0.8 & 75.9 $\pm$ 0.8 & 75.7 $\pm$ 0.8 & 76.1 $\pm$ 0.8 \\
    Fungi & 53.8 $\pm$ 1.2 & 55.3 $\pm$ 1.2 & 56.8 $\pm$ 1.2 & 56.5 $\pm$ 1.2 & 56.9 $\pm$ 1.2 & 57.2 $\pm$ 1.2 \\
    Traffic Sign & 65.9 $\pm$ 1.1 & 66.6 $\pm$ 1.1 & 64.7 $\pm$ 1.1 & 64.7 $\pm$ 1.1 & 65.8 $\pm$ 1.1 & 65.9 $\pm$ 1.1 \\
    MSCOCO & 49.5 $\pm$ 1.1 & 50.0 $\pm$ 1.1 & 48.2 $\pm$ 1.2 & 50.2 $\pm$ 1.1 & 50.0 $\pm$ 1.0 & 51.9 $\pm$ 1.1 \\
    \midrule
    Caltech101 & 87.9  & 87.8  & 87.1  & 87.5  & 87.7  & 88.2\\
    CIFAR100 & 46.8  & 46.6  & 45.2  & 48.1  & 48.8  & 50.1 \\
    Flowers102 & 82.8  & 83.8  & 82.9  & 83.7  & 83.5  & 83.0 \\
    Pets  & 89.2  & 89.2  & 89.3  & 89.5  & 89.3  & 89.7 \\
    Sun397 & 28.8  & 31.5  & 30.3  & 32.4  & 30.9  & 32.3 \\
    SVHN  & 51.4  & 53.0    & 49.6  & 53.5  & 51.0    & 52.7 \\
    \midrule
    EuroSAT & 88.5  & 88.4  & 88.4  & 88.3  & 89.3  & 88.6 \\
    Resics45 & 75.3  & 75.1  & 74.4  & 75.9  & 76.4  & 76.1 \\
    Patch Camelyon & 80.4  & 80.2  & 78.7  & 80.2  & 81.4  & 81.9 \\
    Retinopathy & 42.8  & 42.0    & 40.4  & 40.7  & 40.3  & 39.8 \\
    \midrule
    CLEVR-count & 31.0    & 29.0    & 30.4  & 29.7  & 31.4  & 30.9 \\
    CLEVR-dist & 33.4  & 33.4  & 32.6  & 32.8  & 32.8  & 33.0 \\
    dSprites-loc & 10.7  & 10.5  & 11.6  & 11.3  & 12.3  & 10.6 \\
    dSprites-ori & 30.5  & 29.9  & 29.4  & 29.2  & 31.1  & 27.9 \\
    SmallNORB-azi & 14.6  & 14    & 14.2  & 14.3  & 14.5  & 14.6 \\
    SmallNORB-elev & 21.3  & 21.3  & 20.8  & 20.7  & 21    & 20.9 \\
    DMLab & 40.4  & 40.6  & 40.3  & 38.5  & 39.4  & 38.8 \\
    KITTI-dist & 65.7  & 61.5  & 63.3  & 63.0    & 63.9  & 62.4 \\
    \midrule
    MD-v2 & 72.8  & 73.7  & 73.3  & 73.8  & 73.9  & 74.3 \\
    VTAB (all) & 51.2  & 51.0    & 50.5  & 51.1  & 51.4  & 51.2 \\
    VTAB (natural) & 64.5  & 65.3  & 64.1  & 65.8  & 65.2  & 66.0 \\
    VTAB (specialized) & 71.8  & 71.4  & 70.5  & 71.3  & 71.9  & 71.6 \\
    VTAB (structured) & 31.0    & 30.0    & 30.3  & 29.9  & 30.8  & 29.9 \\
    \bottomrule
    \end{tabular}%
    \end{adjustbox}
    \label{tab:vtab+md_vary_h}%
\end{table}%

%% file: tables/vtab+md_vary_H_protonets.tex
\begin{table}[htbp]
  \centering
  \caption{Classification accuracy results on \vtabmd{} \citep{dumoulin2021comparing} using ProtoNets with varying values of $|\mathcal{H}|$. Image size is 224 x 224 pixels. All figures are percentages and the $\pm$ sign indicates the 95\% confidence interval over tasks. The \textsc{VTAB-v2} results have no confidence interval as the testing protocol requires only a single run over the entire test set.}
    \vspace{0.9em}
    \begin{tabular}{lccccc}
    \toprule
    Dataset      & $|\mathcal{H}|=0$   & $|\mathcal{H}|=10$  & $|\mathcal{H}|=20$  & $|\mathcal{H}|=30$  & $|\mathcal{H}|=40$ \\
    \midrule
    Omniglot & 86.7  $\pm$ 0.8 & 87.7  $\pm$ 0.8 & 88.3  $\pm$ 0.8 & 88.3  $\pm$ 0.7 & 88.3  $\pm$ 0.8 \\
    Aircraft & 83.8  $\pm$ 0.7 & 84.6  $\pm$ 0.7 & 84.1  $\pm$ 0.7 & 85.1  $\pm$ 0.7 & 85.0  $\pm$ 0.7 \\
    Birds & 88.8  $\pm$ 0.6 & 89.4  $\pm$ 0.6 & 89.8  $\pm$ 0.6 & 89.1  $\pm$ 0.7 & 90.2  $\pm$ 0.5 \\
    DTD   & 78.6  $\pm$ 0.6 & 79.7  $\pm$ 0.6 & 80.2  $\pm$ 0.7 & 80.6  $\pm$ 0.7 & 81.4  $\pm$ 0.6 \\
    QuickDraw & 73.5  $\pm$ 0.8 & 75.0  $\pm$ 0.7 & 75.2  $\pm$ 0.8 & 75.6  $\pm$ 0.7 & 76.0  $\pm$ 0.7 \\
    Fungi & 59.4  $\pm$ 1.2 & 58.9  $\pm$ 1.1 & 58.2  $\pm$ 1.2 & 58.0  $\pm$ 1.1 & 57.4  $\pm$ 1.1 \\
    Traffic Sign & 50.0  $\pm$ 1.1 & 52.2  $\pm$ 1.1 & 52.1  $\pm$ 1.0 & 53.1  $\pm$ 1.1 & 53.5  $\pm$ 1.1 \\
    MSCOCO & 47.3  $\pm$ 1.0 & 48.1  $\pm$ 1.0 & 48.1  $\pm$ 1.1 & 50.2  $\pm$ 1.0 & 49.8  $\pm$ 1.1 \\
    \midrule
    Caltech101 & 86.6  & 86.9  & 87.2  & 87.2  & 87.4 \\
    CIFAR100 & 35.5  & 39.6  & 42.0    & 43.4  & 43.1 \\
    Flowers102 & 76.6  & 77.9  & 78.3  & 78.5  & 78.2 \\
    Pets  & 88.5  & 88.4  & 88.7  & 88.7  & 88.6 \\
    Sun397 & 31.5  & 31.8  & 31.1  & 31.8  & 32.9 \\
    SVHN  & 32.0    & 35.6  & 36.4  & 35.3  & 35.2 \\
    \midrule
    EuroSAT & 79.4  & 81.6  & 81.8  & 82.9  & 83.3 \\
    Resics45 & 65.7  & 67.4  & 68.0    & 69.0    & 68.8 \\
    Patch Camelyon & 75.9  & 72.8  & 73.7  & 74.2  & 73.3 \\
    Retinopathy & 32.9  & 33.7  & 33.2  & 31.9  & 31.3 \\
    CLEVR-count & 28.3  & 27.3  & 27.1  & 27.2  & 27.2 \\
    CLEVR-dist & 29.5  & 29.2  & 29.0    & 28.9  & 28.5 \\
    dSprites-loc & 13.3  & 14.1  & 14.0    & 13.2  & 13.4 \\
    dSprites-ori & 20.4  & 19.6  & 20.4  & 19.8  & 19.6 \\
    SmallNORB-azi & 9.4   & 9.4   & 9.6   & 9.5   & 9.4 \\
    SmallNORB-elev & 16.3  & 17.1  & 17.0    & 17.1  & 17.0 \\
    DMLab & 35.2  & 35.5  & 35.9  & 35.9  & 35.8 \\
    KITTI-dist & 55.6  & 57.2  & 58.2  & 57.1  & 56.5 \\
    \midrule
    MD-v2 & 71.0    & 72.0    & 72.0    & 72.5  & 72.7 \\
    VTAB (all) & 45.1  & 45.8  & 46.2  & 46.2  & 46.1 \\
    VTAB (natural) & 58.5  & 60.0    & 60.6  & 60.8  & 60.9 \\
    VTAB (specialized) & 63.5  & 63.9  & 64.2  & 64.5  & 64.2 \\
    VTAB (structured) & 26.0    & 26.2  & 26.4  & 26.1  & 25.9 \\
    \bottomrule
    \end{tabular}%
  \label{tab:vtab+md_vary_h_protonets}%
\end{table}%

%% file: tables/vtab+md_vary_H_84.tex
\begin{table}[htbp]
  \centering
  \caption{Classification accuracy results on \vtabmd{} \citep{dumoulin2021comparing} using \scl{} with two values of $|\mathcal{H}|$. Image size is 84 x 84 pixels. All figures are percentages and the $\pm$ sign indicates the 95\% confidence interval over tasks. The \textsc{VTAB-v2} results have no confidence interval as the testing protocol requires only a single run over the entire test set.}
    \vspace{0.9em}
    \begin{tabular}{lcc}
    \toprule
    Dataset & $|\mathcal{H}|=40$  & $|\mathcal{H}|=\mathcal|{D}_S|$ \\
    \midrule
    Omniglot & 83.7$\pm$1.0 & 90.9$\pm$0.6 \\
    Aircraft & 65.4$\pm$0.9 & 77.5$\pm$0.7 \\
    Birds & 69.5$\pm$1.0 & 76.4$\pm$0.8 \\
    DTD   & 72.1$\pm$0.8 & 74.3$\pm$0.7 \\
    QuickDraw & 70.6$\pm$0.9 & 76.5$\pm$0.7 \\
    Fungi & 45.5$\pm$1.2 & 51.3$\pm$1.1 \\
    Traffic Sign & 58.2$\pm$1.0 & 54.8$\pm$1.1 \\
    MSCOCO & 43.8$\pm$1.1 & 45.1$\pm$1.0 \\
    \midrule
    Caltech101 & 74.9  & 79.6 \\
    CIFAR100 & 35.4  & 37.1 \\
    Flowers102 & 69.6  & 65.5 \\
    Pets  & 55.5  & 69.8 \\
    Sun397 & 13.9  & 18.0 \\
    SVHN  & 36.7  & 26.7 \\
    \midrule
    EuroSAT & 84.2  & 82.8 \\
    Resics45 & 61.5  & 64.5 \\
    Patch Camelyon & 74.0    & 78.4 \\
    Retinopathy & 24.3  & 29.4 \\
    CLEVR-count & 32.2  & 30.7 \\
    CLEVR-dist & 36.3  & 32.5 \\
    dSprites-loc & 26.5  & 43.9 \\
    dSprites-ori & 19.7  & 21.1 \\
    SmallNORB-azi & 14.0    & 13.5 \\
    SmallNORB-elev & 19.0    & 19.6 \\
    DMLab & 33.4  & 33.9 \\
    KITTI-dist & 58.1  & 58.1 \\
    \midrule
    MD-v2 & 63.6  & 68.4 \\
    VTAB (all) & 42.7  & 44.7 \\
    VTAB (natural) & 47.7  & 49.5 \\
    VTAB (specialized) & 61.0    & 63.8 \\
    VTAB (structured) & 29.9  & 31.7 \\
    \bottomrule
    \end{tabular}%
  \label{tab:vtab+md_vary_h_84}%
\end{table}%

%% file: tables/mean_std_vs_h.tex
\begin{table}[htbp]
  \centering
  \caption{Mean Squared Error (lower is better) between the mean of the gradient estimates and the true gradients for both \lite{} and subsampled small tasks as $|\mathcal{H}|$ is varied. The task used was a 10-way, 10-shot task of $84 \times 84$ pixels from the DTD dataset. For each value of $|\mathcal{H}|$, 1000 support set examples were used.}
    \vspace{0.9em}
    \begin{adjustbox}{max width=\textwidth}
    \begin{tabular}{l|ccccccccc}
    \toprule
          & \multicolumn{9}{c}{$|\mathcal{H}|$} \\
    Training Mode   & 10    & 20    & 30    & 40    & 50    & 60    & 70    & 80    & 90 \\
    \midrule
    LITE  & 9.53E-11 & 9.24E-11 & 7.89E-11 & 8.48E-11 & 5.11E-11 & 5.31E-11 & 6.03E-11 & 1.02E-10 & 2.51E-11 \\
    Subsampled Small Task & 9.23E-11 & 8.46E-11 & 7.67E-11 & 7.15E-11 & 6.45E-11 & 6.27E-11 & 5.67E-11 & 4.78E-11 & 4.30E-11 \\
    \bottomrule
    \end{tabular}%
    \end{adjustbox}
  \label{tab:mse_vs_h}%
\end{table}%

\begin{table}[htbp]
  \centering
  \caption{Average root mean squared error (lower is better) with respect to the exact gradients for both \lite{} and subsampled small tasks as $|\mathcal{H}|$ is varied. The task used was a 10-way, 10-shot task of $84 \times 84$ pixels from the DTD dataset. For each value of $|\mathcal{H}|$, 1000 support set examples were used.}
  \vspace{0.9em}
    \begin{adjustbox}{max width=\textwidth}
    \begin{tabular}{l|ccccccccc}
    \toprule
          & \multicolumn{9}{c}{$|\mathcal{H}|$} \\
    Training Mode   & 10    & 20    & 30    & 40    & 50    & 60    & 70    & 80    & 90 \\
    \midrule
    LITE  & 4.35E-03 & 3.20E-03 & 2.55E-03 & 2.32E-03 & 1.84E-03 & 1.63E-03 & 1.65E-03 & 1.73E-03 & 1.06E-03 \\
    Subsampled Small Task & 5.56E-03 & 4.32E-03 & 3.43E-03 & 2.66E-03 & 2.37E-03 & 2.04E-03 & 1.77E-03 & 1.40E-03 & 1.13E-03 \\
    \bottomrule
    \end{tabular}%
    \end{adjustbox}
  \label{tab:std_vs_h}%
\end{table}%

%% file: tables/vtab+md_330.tex
\begin{table}[htbp]
  \centering
  \caption{Classification accuracy results on \vtabmd{} \citep{dumoulin2021comparing} using \scl{} with $|\mathcal{H}|=10$ and image size is $320 \times 320$ pixels. All figures are percentages and the $\pm$ sign indicates the 95\% confidence interval over tasks. The \textsc{VTAB-v2} results have no confidence interval as the testing protocol requires only a single run over the entire test set.}
    \vspace{0.9em}
    \begin{tabular}{lc}
    \toprule
    Dataset & $|\mathcal{H}|=10$, 320 x 320 pixels \\
    \midrule
    Omniglot & 83.2$\pm$1.0 \\
    Aircraft & 82.5$\pm$0.8 \\
    Birds & 91.2$\pm$0.6 \\
    DTD   & 85.3$\pm$0.7 \\
    QuickDraw & 74.1$\pm$0.8 \\
    Fungi & 58.0$\pm$1.2 \\
    Traffic Sign & 62.4$\pm$1.1 \\
    MSCOCO & 46.8$\pm$1.1 \\
    \midrule
    Caltech101 & 88.0 \\
    CIFAR100 & 45.2 \\
    Flowers102 & 82.6 \\
    Pets  & 89.5 \\
    Sun397 & 28.6 \\
    SVHN  & 51.9 \\
    \midrule
    EuroSAT & 86.3 \\
    Resics45 & 72.7 \\
    Patch Camelyon & 80.7 \\
    Retinopathy & 46.4 \\
    \midrule
    CLEVR-count & 30.8 \\
    CLEVR-dist & 33.5 \\
    dSprites-loc & 14.2 \\
    dSprites-ori & 28.2 \\
    SmallNORB-azi & 14.0 \\
    SmallNORB-elev & 20.7 \\
    DMLab & 40.3 \\
    KITTI-dist & 62.3 \\
    \midrule
    MD-v2 & 72.9 \\
    VTAB (all) & 50.9 \\
    VTAB (natural) & 64.3 \\
    VTAB (specialized) & 71.5 \\
    VTAB (structured) & 30.5 \\
    \bottomrule
    \end{tabular}%
  \label{tab:vtab+md_vary_330}%
\end{table}%

%% file: main.bbl
\begin{thebibliography}{36}
\providecommand{\natexlab}[1]{#1}
\providecommand{\url}[1]{\texttt{#1}}
\expandafter\ifx\csname urlstyle\endcsname\relax
  \providecommand{\doi}[1]{doi: #1}\else
  \providecommand{\doi}{doi: \begingroup \urlstyle{rm}\Url}\fi

\bibitem[Finn et~al.(2017)Finn, Abbeel, and Levine]{finn2017model}
Chelsea Finn, Pieter Abbeel, and Sergey Levine.
\newblock Model-agnostic meta-learning for fast adaptation of deep networks.
\newblock In \emph{\icml{34th}}, pages 1126--1135, 2017.

\bibitem[Zintgraf et~al.(2019)Zintgraf, Shiarlis, Kurin, Hofmann, and
  Whiteson]{zintgraf2018cavia}
Luisa~M. Zintgraf, Kyriacos Shiarlis, Vitaly Kurin, Katja Hofmann, and Shimon
  Whiteson.
\newblock Fast context adaptation via meta-learning.
\newblock In \emph{\icml{36th}}, 2019.

\bibitem[Snell et~al.(2017)Snell, Swersky, and Zemel]{snell2017prototypical}
Jake Snell, Kevin Swersky, and Richard Zemel.
\newblock Prototypical networks for few-shot learning.
\newblock In \emph{\neurips{31st}}, pages 4077--4087, 2017.

\bibitem[Requeima et~al.(2019)Requeima, Gordon, Bronskill, Nowozin, and
  Turner]{requeima2019cnaps}
James Requeima, Jonathan Gordon, John Bronskill, Sebastian Nowozin, and
  Richard~E Turner.
\newblock Fast and flexible multi-task classification using conditional neural
  adaptive processes.
\newblock In \emph{\neurips{33rd}}, pages 7957--7968, 2019.

\bibitem[Bateni et~al.(2020)Bateni, Goyal, Masrani, Wood, and
  Sigal]{bateni2020improved}
Peyman Bateni, Raghav Goyal, Vaden Masrani, Frank Wood, and Leonid Sigal.
\newblock Improved few-shot visual classification.
\newblock In \emph{\cvpr}, pages 14493--14502, 2020.

\bibitem[Kolesnikov et~al.(2019)Kolesnikov, Beyer, Zhai, Puigcerver, Yung,
  Gelly, and Houlsby]{kolesnikov2019big}
Alexander Kolesnikov, Lucas Beyer, Xiaohua Zhai, Joan Puigcerver, Jessica Yung,
  Sylvain Gelly, and Neil Houlsby.
\newblock Big transfer (bit): General visual representation learning.
\newblock \emph{arXiv preprint arXiv:1912.11370}, 6\penalty0 (2):\penalty0 8,
  2019.

\bibitem[Ring(1997)]{ring1997child}
Mark~B Ring.
\newblock {CHILD: A first step towards continual learning}.
\newblock \emph{Machine Learning}, 28\penalty0 (1):\penalty0 77--104, 1997.

\bibitem[Saad(2009)]{saad2009line}
David Saad.
\newblock \emph{On-line learning in neural networks}.
\newblock Number~17. Cambridge University Press, 2009.

\bibitem[Hoi et~al.(2018)Hoi, Sahoo, Lu, and Zhao]{hoi2018online}
Steven~CH Hoi, Doyen Sahoo, Jing Lu, and Peilin Zhao.
\newblock Online learning: A comprehensive survey.
\newblock \emph{arXiv preprint arXiv:1802.02871}, 2018.

\bibitem[Antoniou et~al.(2020)Antoniou, Patacchiola, Ochal, and
  Storkey]{antoniou2020defining}
Antreas Antoniou, Massimiliano Patacchiola, Mateusz Ochal, and Amos Storkey.
\newblock Defining benchmarks for continual few-shot learning.
\newblock \emph{arXiv preprint arXiv:2004.11967}, 2020.

\bibitem[Dumoulin et~al.(2021)Dumoulin, Houlsby, Evci, Zhai, Goroshin, Gelly,
  and Larochelle]{dumoulin2021comparing}
Vincent Dumoulin, Neil Houlsby, Utku Evci, Xiaohua Zhai, Ross Goroshin, Sylvain
  Gelly, and Hugo Larochelle.
\newblock Comparing transfer and meta learning approaches on a unified few-shot
  classification benchmark.
\newblock \emph{arXiv preprint arXiv:2104.02638}, 2021.

\bibitem[Chen et~al.(2016)Chen, Xu, Zhang, and Guestrin]{chen2016training}
Tianqi Chen, Bing Xu, Chiyuan Zhang, and Carlos Guestrin.
\newblock Training deep nets with sublinear memory cost.
\newblock \emph{arXiv preprint arXiv:1604.06174}, 2016.

\bibitem[Triantafillou et~al.(2020)Triantafillou, Zhu, Dumoulin, Lamblin, Evci,
  Xu, Goroshin, Gelada, Swersky, Manzagol, and
  Larochelle]{triantafillou2019meta}
Eleni Triantafillou, Tyler Zhu, Vincent Dumoulin, Pascal Lamblin, Utku Evci,
  Kelvin Xu, Ross Goroshin, Carles Gelada, Kevin Swersky, Pierre-Antoine
  Manzagol, and Hugo Larochelle.
\newblock {Meta-Dataset: A Dataset of Datasets for Learning to Learn from Few
  Examples}.
\newblock In \emph{\iclr{8th}}, 2020.

\bibitem[Massiceti et~al.(2021)Massiceti, Zintgraf, Bronskill, Theodorou,
  Harris, Cutrell, Morrison, Hofmann, and Stumpf]{massiceti2021orbit}
Daniela Massiceti, Luisa Zintgraf, John Bronskill, Lida Theodorou,
  Matthew~Tobias Harris, Edward Cutrell, Cecily Morrison, Katja Hofmann, and
  Simone Stumpf.
\newblock {ORBIT: A Real-World Few-Shot Dataset for Teachable Object
  Recognition}.
\newblock In \emph{\iccv}, 2021.

\bibitem[Vinyals et~al.(2016)Vinyals, Blundell, Lillicrap, Kavukcuoglu, and
  Wierstra]{vinyals2016matching}
Oriol Vinyals, Charles Blundell, Timothy Lillicrap, Koray Kavukcuoglu, and Daan
  Wierstra.
\newblock Matching networks for one shot learning.
\newblock In \emph{\neurips{30th}}, pages 3630--3638, 2016.

\bibitem[Ravi and Larochelle(2017)]{ravi2016optimization}
Sachin Ravi and Hugo Larochelle.
\newblock Optimization as a model for few-shot learning.
\newblock In \emph{\iclr{5th}}, 2017.

\bibitem[Gordon et~al.(2019)Gordon, Bronskill, Bauer, Nowozin, and
  Turner]{gordon2018meta}
Jonathan Gordon, John Bronskill, Matthias Bauer, Sebastian Nowozin, and Richard
  Turner.
\newblock Meta-learning probabilistic inference for prediction.
\newblock In \emph{\iclr{7th}}, 2019.

\bibitem[Perez et~al.(2018)Perez, Strub, De~Vries, Dumoulin, and
  Courville]{perez2018film}
Ethan Perez, Florian Strub, Harm De~Vries, Vincent Dumoulin, and Aaron
  Courville.
\newblock Fi{LM}: Visual reasoning with a general conditioning layer.
\newblock In \emph{\aaai{32nd}}, 2018.

\bibitem[Mahalanobis(1936)]{mahalanobis1936generalized}
Prasanta~Chandra Mahalanobis.
\newblock On the generalized distance in statistics.
\newblock In \emph{Proceedings of the National Institute of Science of India},
  volume~12, pages 49--55, 1936.

\bibitem[Zhai et~al.(2019)Zhai, Puigcerver, Kolesnikov, Ruyssen, Riquelme,
  Lucic, Djolonga, Pinto, Neumann, Dosovitskiy, et~al.]{zhai2019large}
Xiaohua Zhai, Joan Puigcerver, Alexander Kolesnikov, Pierre Ruyssen, Carlos
  Riquelme, Mario Lucic, Josip Djolonga, Andre~Susano Pinto, Maxim Neumann,
  Alexey Dosovitskiy, et~al.
\newblock A large-scale study of representation learning with the visual task
  adaptation benchmark.
\newblock \emph{arXiv preprint arXiv:1910.04867}, 2019.

\bibitem[Sun et~al.(2017)Sun, Shrivastava, Singh, and Gupta]{sun2017revisiting}
Chen Sun, Abhinav Shrivastava, Saurabh Singh, and Abhinav Gupta.
\newblock Revisiting unreasonable effectiveness of data in deep learning era.
\newblock In \emph{\iccv}, pages 843--852, 2017.

\bibitem[Dvornik et~al.(2020)Dvornik, Schmid, and Mairal]{dvornik2020selecting}
Nikita Dvornik, Cordelia Schmid, and Julien Mairal.
\newblock Selecting relevant features from a multi-domain representation for
  few-shot classification.
\newblock In \emph{\eccv}, pages 769--786, 2020.

\bibitem[Hospedales et~al.(2020)Hospedales, Antoniou, Micaelli, and
  Storkey]{hospedales2020meta}
Timothy Hospedales, Antreas Antoniou, Paul Micaelli, and Amos Storkey.
\newblock Meta-learning in neural networks: A survey.
\newblock \emph{arXiv preprint arXiv:2004.05439}, 2020.

\bibitem[Doersch et~al.(2020)Doersch, Gupta, and
  Zisserman]{doersch2020crosstransformers}
Carl Doersch, Ankush Gupta, and Andrew Zisserman.
\newblock {CrossTransformers: spatially-aware few-shot transfer}.
\newblock \emph{arXiv preprint arXiv:2007.11498}, 2020.

\bibitem[Nichol et~al.(2018)Nichol, Achiam, and Schulman]{nichol2018first}
Alex Nichol, Joshua Achiam, and John Schulman.
\newblock On first-order meta-learning algorithms.
\newblock \emph{arXiv preprint arXiv:1803.02999}, 2018.

\bibitem[Rajeswaran et~al.(2019)Rajeswaran, Finn, Kakade, and
  Levine]{rajeswaran2019meta}
Aravind Rajeswaran, Chelsea Finn, Sham Kakade, and Sergey Levine.
\newblock Meta-learning with implicit gradients.
\newblock In \emph{\neurips}, 2019.

\bibitem[Shin et~al.(2021)Shin, Lee, Gong, and Hwang]{shin2021large}
Jaewoong Shin, Hae~Beom Lee, Boqing Gong, and Sung~Ju Hwang.
\newblock Large-scale meta-learning with continual trajectory shifting.
\newblock In \emph{\icml}, 2021.

\bibitem[Yosinski et~al.(2014)Yosinski, Clune, Bengio, and
  Lipson]{yosinski2014transferable}
Jason Yosinski, Jeff Clune, Yoshua Bengio, and Hod Lipson.
\newblock How transferable are features in deep neural networks?
\newblock In \emph{\neurips{28th}}, pages 3320--3328, 2014.

\bibitem[Deng et~al.(2009)Deng, Dong, Socher, Li, Li, and
  Fei-Fei]{deng2009imagenet}
Jia Deng, Wei Dong, Richard Socher, Li-Jia Li, Kai Li, and Li~Fei-Fei.
\newblock {ImageNet}: A large-scale hierarchical image database.
\newblock In \emph{\cvpr}, 2009.

\bibitem[Zakharov et~al.(2019)Zakharov, Shysheya, Burkov, and
  Lempitsky]{zakharov2019few}
Egor Zakharov, Aliaksandra Shysheya, Egor Burkov, and Victor Lempitsky.
\newblock Few-shot adversarial learning of realistic neural talking head
  models.
\newblock In \emph{\iccv}, pages 9459--9468, 2019.

\bibitem[He et~al.(2016)He, Zhang, Ren, and Sun]{he2016deep}
Kaiming He, Xiangyu Zhang, Shaoqing Ren, and Jian Sun.
\newblock Deep residual learning for image recognition.
\newblock In \emph{\cvpr}, pages 770--778, 2016.

\bibitem[Tan and Le(2019)]{tan2019efficientnet}
Mingxing Tan and Quoc Le.
\newblock {EfficientNet: Rethinking model scaling for convolutional neural
  networks}.
\newblock In \emph{\icml{36th}}, pages 6105--6114, 2019.

\bibitem[Sandler et~al.(2018)Sandler, Howard, Zhu, Zhmoginov, and
  Chen]{sandler2018mobilenetv2}
Mark Sandler, Andrew Howard, Menglong Zhu, Andrey Zhmoginov, and Liang-Chieh
  Chen.
\newblock Mobilenetv2: Inverted residuals and linear bottlenecks.
\newblock In \emph{\cvpr}, pages 4510--4520, 2018.

\bibitem[Russakovsky et~al.(2015)Russakovsky, Deng, Su, Krause, Satheesh, Ma,
  Huang, Karpathy, Khosla, Bernstein, Berg, and Fei-Fei]{ILSVRC15}
Olga Russakovsky, Jia Deng, Hao Su, Jonathan Krause, Sanjeev Satheesh, Sean Ma,
  Zhiheng Huang, Andrej Karpathy, Aditya Khosla, Michael Bernstein,
  Alexander~C. Berg, and Li~Fei-Fei.
\newblock {ImageNet Large Scale Visual Recognition Challenge}.
\newblock \emph{\ijcv}, 115\penalty0 (3):\penalty0 211--252, 2015.
\newblock \doi{10.1007/s11263-015-0816-y}.

\bibitem[Kingma and Ba(2015)]{kingma2014adam}
Diederik Kingma and Jimmy Ba.
\newblock Adam: A method for stochastic optimization.
\newblock In \emph{\iclr{3rd}}, 2015.

\bibitem[TFD(2020)]{TFDS}
{TensorFlow Datasets}, a collection of ready-to-use datasets.
\newblock \url{https://www.tensorflow.org/datasets}, 2020.

\end{thebibliography}
